\documentclass[journal]{IEEEtran}
\IEEEoverridecommandlockouts
\usepackage{cite}
\usepackage{amsmath,amsthm,amssymb,amsfonts}
\usepackage{array}
\usepackage{graphicx}
\usepackage{textcomp}
\usepackage[mathscr]{eucal}
\usepackage[table]{xcolor}
\usepackage{xcolor}
\usepackage{cleveref}
\crefformat{section}{\S#2#1#3}
\usepackage[framemethod=tikz]{mdframed}
\usepackage[ruled,vlined,linesnumbered]{algorithm2e}
\usepackage{algpseudocode}
\usepackage{multirow}
\usepackage{rotating}

\usepackage[font=footnotesize]{caption}
\usepackage{subcaption}
\usepackage{pifont}
\usepackage[flushmargin]{footmisc}
\newcommand{\xmark}{\ding{55}}
\newcommand{\cmark}{\ding{51}}

\algnewcommand\LeftComment[2]{%
\hspace{#1\algindent}$\triangleright$ \eqparbox{COMMENT}{#2} \hfill %
}

\let\oldnl\nl% Store \nl in \oldnl
\newcommand{\nonl}{\renewcommand{\nl}{\let\nl\oldnl}}

\def\BibTeX{{\rm B\kern-.05em{\sc i\kern-.025em b}\kern-.08em
    T\kern-.1667em\lower.7ex\hbox{E}\kern-.125emX}}
\begin{document}

%\title{Deep Leanring Compression Techniques:}
\title{A Survey on Deep Neural Network Compression: Challenges, Overview, and Solutions}
\author{\IEEEauthorblockN{Rahul Mishra, Hari Prabhat Gupta, and Tanima Dutta} \vspace{-0.5cm}
 \thanks{The authors are with the Department of Computer Science and Engineering, Indian Institute of Technology (BHU) Varanasi, India (e-mail: rahulmishra.rs.cse17@iitbhu.ac.in;  hariprabhat.cse@iitbhu.ac.in; tanima.cse@iitbhu.ac.in)}}

\maketitle
\begin{abstract} 
Deep Neural Network (DNN) has gained unprecedented performance due to its automated feature extraction capability. This high order performance leads to significant incorporation of DNN models in different Internet of Things (IoT) applications in the past decade. However, the colossal requirement of computation, energy, and storage of DNN models make their deployment prohibitive on resource constraint IoT devices. Therefore, several compression techniques were proposed in recent years for reducing the storage and computation requirements of the DNN model. These techniques on DNN compression have utilized a different perspective for compressing DNN with minimal accuracy compromise. It encourages us to make a comprehensive overview of the DNN compression techniques. In this paper, we present a comprehensive review of existing literature on compressing DNN model that reduces both storage and computation requirements. We divide the existing approaches into five broad categories, \textit{i.e.,} network pruning, sparse representation, bits precision, knowledge distillation, and miscellaneous, based upon the mechanism incorporated for compressing the DNN model. The paper also discussed the challenges associated with each category of DNN compression techniques. Finally, we provide a quick summary of existing work under each category with the future direction in DNN compression.
\end{abstract}


\section{Introduction}
Recent years have witnessed significant growth in machine learning-based Internet of Things (IoT) applications due to easier and cheaper availability of small computing devices~\cite{hussain2020machine,8936464}. The machine learning-based approaches require handcrafted features which reduces its suitability for a new dataset, where, information about data distribution is missing~\cite{xie2018survey}. Therefore, Deep Neural Network (DNN) based IoT applications have achieved dominance in different application domains, including smart home, agriculture, locomotion mode recognition, \textit{etc}~\cite{yao2018deep, 9164991,9209121,9166711,9136920,8998403}. It is because the DNN provides automatic feature extraction capability from a large amount of raw data.  The DNN models not only eliminates the human intervention for calculating domain-related features but achieves coercive performance~\cite{yao2018deep}. DNN model involves the following components, including Convolutional Neural Network (CNN), Fully Connected (FC) layers, and RNN (Recurrent Neural network). A DNN model can have all these components at a time or any of its combination. The CNN extract spatial features and RNN identifies temporal features form the dataset. These features archives a high order performance in classification, regression and prediction of different tasks~\cite{tu2019deep,8936677}.

Despite these advantages, the DNN models require a significant amount of resources including, energy, processing capacity, and storage. This resource requirement reduces the suitability of DNN model for Resources constraint Devices (RCDs)~\cite{126846}. The huge resource requirements of the DNN model also become a bottleneck for real-time inference and to run DNN model on browser-based applications. Therefore, to mitigate these shortcomings of DNN model, \textit{i.e.,} energy, processing capacity, and storage, different DNN compression techniques have been proposed in the existing literature.

We can achieve several benefits by compressing the DNN model over the traditional cumbersome DNN model. Some of these benefits are as follows:

\begin{itemize}
 \item \textit{Storage capacity:} The DNN model achieves significant accuracy that comes up with a large number of parameters, which requires considerable storage~\cite{ke2018nnest, palit2019uniform}. Therefore, by compressing the DNN model, we can preserve storage that facilitates the deployment of DNN model on Resource Constraint Devices (RCDs).
 
 \item \textit{Computation requirements:} A large number of FLoating point OPerations (FLOPs) involve in the DNN operation can exceed the limited computational capacity of the RCDs~\cite{marco2020optimizing}. Thus, it would be beneficial to employ DNN compression for reducing the computation requirement.

 \item \textit{Earliness:} The training and inference time of a DNN model is significantly high that hampers its real-time inference performance~\cite{xiang2019pipelined, 8949724}. The DNN compression mechanisms, therefore, provides a higher degree of earliness in both training and inference phases.
 
 \item \textit{Privacy:} The data transmission from the source to the high-end machine leads to the security breach and privacy compromise~\cite{akmandor2018smart}. Thus, it would be beneficial to employ in-situ processing using compressed DNN model on RCDs, which helps in preserving privacy and provides data security.
 
 \item \textit{Energy consumption:} The DNN compression also preserves energy for processing the data~\cite{yang2017method, wu2019accelergy, 8811605}. It enhances its suitability for the deployment of compressed DNN model on battery-operated IoT devices.
\end{itemize}

The demand for IoT based applications is growing day-by-day that encourages data scientists to incorporate DNN models in IoT applications. Therefore, it is beneficial to provide a thorough overview of the DNN compression technique that can meet out the limited storage and processing capacity available at the resource constraint IoT devices. Thus, we carry out a comprehensive survey of existing literature on different DNN compression technique and propose a useful categorization to point out a potential research gap for future work. This paper presents a systematic overview of DNN compression techniques that reduce both storage and computation requirements of the DNN model. We categorize the different DNN compression techniques into five broad categories based on the type of strategy they followed for compression DNN model with minimal accuracy compromise. The five broad categories for DNN compression are network pruning, sparse representation, bits precision, knowledge distillation, and miscellaneous.  Fig.~\ref{category} illustrates the categorization hierarchy with different DNN compression techniques and their subcategories.

Next, section discusses the categorization of the DNN compression technique. Section~\ref{np} presents the overview of network pruning technique with all sub-categories. The summary of the sparse representation technique is presented in Section~\ref{sr}. Further, Section~\ref{bp} and section~\ref{kd} demonstrate the detailed description of bits precision and knowledge distillation techniques, respectively. The miscellaneous techniques that are not included in the above four category is discussed in Section~\ref{mln}. Finally, Section~\ref{fd} presents the discussions and future directions of the DNN compression techniques.

\section{Categorization of DNN compression techniques}
\begin{figure*}[ht]
 \centering
 \includegraphics[scale=1.0]{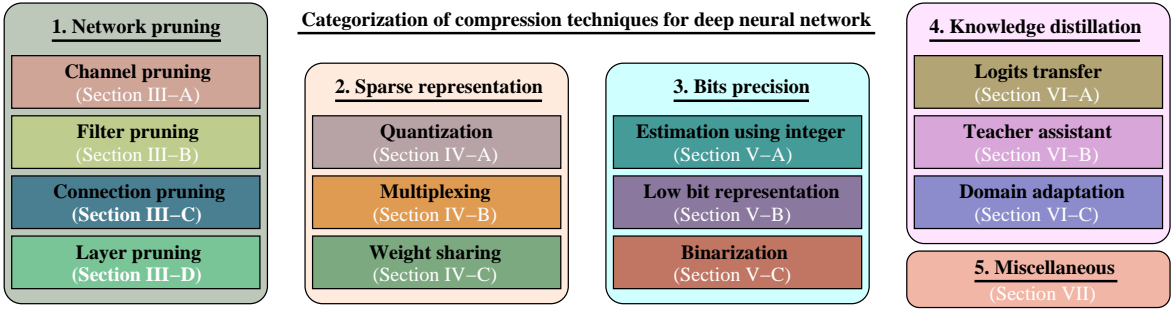}
 \caption{Overview of different categories of compression techniques for deep neural network.}
 \label{category}
\end{figure*}

This section classifies the existing work on DNN compression in five broad categories, \textit{i.e.,} network pruning, sparse representation, bits precision, knowledge distillation, and miscellaneous techniques. These categories were chosen through an extensive literature review on DNN compression techniques. For example, based on the network shrinking, we come up with network pruning, and through sparsification of the weight matrix, we can obtain the category for sparse representation. Fig.~\ref{category} illustrates the overview of the paper with different categories of DNN compression.  Table~\ref{chal1} summarises the existing work on DNN compression under different categories. The broad categories of the DNN compression techniques are defined as

\subsection{Network pruning}
Deep network pruning is one of the popular techniques to reduce the size of a deep learning model by incorporating the removal of the inadequate components such as channels, filters, neurons, or layers to produce a lightweight model~\cite{10.1145/3005348,blalock2020state}. The resultant lightweight model is a low power consuming, memory-efficient, and provides faster inference with minimal accuracy compromise. The adequacy of a component relies on the amount of loss incurred when a component is removed from the model~\cite{molchanov2019importance}. Sometimes pruning is also said to be a binary criterion to determine the removal or persistence of a component in the DNN. The pruning is performed on a pre-trained model iteratively, such that only inadequate components are pruned from the model. We further categorized the network pruning techniques into four parts, \textit{i.e.,} channel pruning, filter pruning, connection pruning, and layer pruning. It helps in reducing the storage and computation requirements of the DNN model. The detailed overview of existing network pruning techniques~\cite{he2017channel,liu2018improvement,liu2017learning, yan2019vargfacenet,howard2017mobilenets, luo2017thinet, bhattacharya2016sparsification, zhou2019knee, huynh2017deepmon,denton2014exploiting, han2015deep,yao2017deepiot,han2016eie,qin2018compress,parashar2017scnn,molchanov2017variational,louizos2017bayesian,guo2016dynamic, he2018multi,tan2019mnasnet, chauhan2018performance} for the DNN model is discussed in Section~\ref{np}.

\subsection{Sparse representation}
Sparse representation exploits the sparsity present in the weight matrices of the DNN model~\cite{guo2018sparse}. In sparse representation, the weights there are zero or near to zero are removed from the weight matrix, which reduces the storage and computation requirements of the DNN model. In other words, those links in the network having similar weights are multiplexed, where, in-place of \textit{multiple weights with a single link} is replaced with a \textit{single weight with the multiplex link}. The sparse representation includes low-rank estimates~\cite{zhao2016low}, quantization, multiplexing, \textit{etc.} The principle motive behind the sparse representation is to concise the weight matrix without losing the performance of the DNN model. In this work, we illustrate the overview of existing approaches on DNN compression that incorporates sparse representation. Further, we categorized the DNN  compression using sparse representation into three sub-categories, \textit{i.e.,} quantization, multiplexing, and weight sharing. Section~\ref{sr}presents a detailed description of the sparse representation techniques in the existing literature~\cite{ han2015deep,qin2018compress,jacob2018quantization ,chauhan2018performance, han2016eie, eshratifar2020run, zhang2019mbfn,he2018multi,wu2018sharing,georgiev2017low} on DNN compression.
 
\subsection{Bits precision}
Bits precision in the DNN compression technique reduces the number of bits required for storing the weights in the weight matrices $\mathbf{W}$. For example, the FLOPs in the DNN model requires $32$ bits, which can be replaced with integer operation that requires only $8$ bits. Similarly, we can use binary, $6$ bits, $16$ bits for replacing $32$ bits FLOPs in DNN model~\cite{oh2018portable}. We categorized the existing literature on DNN compression using bits precision into three sub-categories, namely, estimation using integer, low bits representation, and binarization. Section~\ref{bp} highlights the overview of different bits precision techniques~\cite{jacob2018quantization,lee2019neuro, hubara2017quantized, hubara2016binarized,yang2019cdeeparch,umuroglu2017finn,louizos2017bayesian} for DNN compression using limited  number of bits.
 
 \subsection{Knowledge distillation}
In DNN model, the term knowledge distillation is defined as the process of transferring the generalization ability of the cumbersome model (teacher) to the compact model (student) to improve its performance~\cite{liu2018improving}. Knowledge distillation provides a mechanism to overcome the accuracy compromise due to DNN compression. The training of the student model using knowledge distillation improves its generalization ability such that it can mimic similar behaviour as the teacher model for predicting the class label probabilities. In this work, we have covered a detailed overview of three categories of knowledge distillation, \textit{i.e,} logits transfer, teacher assistant, and domain adaptation. Section~\ref{kd} provides a detailed description of the all subcategories of knowledge distillation covered in the existing literature~\cite{wang2019private,hinton2015distilling,yun2020regularizing,li2020few,cheng2020explaining, mirzadeh2019improved,duong2019shrinkteanet,yang2020mobileda,9203948}.
 
\subsection{Miscellaneous}
The DNN compression technique that do-not qualifies the above four categories are classified as miscellaneous. They include such DNN compression techniques, which perform modelling of DNN in such a manner that can be easily fitted on mobile devices. They adopt parallelization of features or task distribution to reduce the storage and computation requirements of the DNN model. The existing literature~\cite{yao2017deepsense, yao2018fastdeepiot, abc,radu2016towards,fang2018nestdnn,radu2018multimodal,mathur2017deepeye,xu2018deepcache,shen2019deepapp,xue2019deepfusion,svyatkovskiy2017training,wang2018deep,zhang2018shufflenet,9130098} on DNN compression under miscellaneous category are discussed in Section~\ref{mln}.
 
\begin{table*}[htbp]
\centering
\caption{Categorization of existing literature on DNN compression techniques.}
\label{chal1}
\begin{tabular}{|l|ll}
\hline
\multirow{17}{*}{\textbf{\begin{tabular}[c]{@{}c@{}}DNN compression\\ techniques\end{tabular}}} 

& \multicolumn{1}{l|}{\multirow{4}{*}{Network pruning }} 
& \multicolumn{1}{l|}{Channel pruning ~\cite{he2017channel,liu2018improvement,chen2019evolutionary,liu2017learning, yan2019vargfacenet,howard2017mobilenets}} \\ \cline{3-3} 
& \multicolumn{1}{l|}{}& \multicolumn{1}{l|}{Filter pruning~\cite{luo2017thinet, bhattacharya2016sparsification, zhou2019knee, huynh2017deepmon,denton2014exploiting}}                                                  \\ \cline{3-3} 
& \multicolumn{1}{l|}{}& \multicolumn{1}{l|}{Connection pruning~\cite{han2015deep,yao2017deepiot,han2016eie,he2018multi,chen2019evolutionary,qin2018compress, parashar2017scnn,molchanov2017variational,louizos2017bayesian,guo2016dynamic}}                                            \\ \cline{3-3} 
& \multicolumn{1}{l|}{}& \multicolumn{1}{l|}{Layer pruning~\cite{he2018multi,tan2019mnasnet, chauhan2018performance}}                                                  \\ \cline{2-3} 
&                                                                              &                                                                                                       \\ \cline{2-3}

& \multicolumn{1}{l|}{\multirow{3}{*}{Sparse representation}}
& \multicolumn{1}{l|}{Quantization ~\cite{han2015deep,qin2018compress,jacob2018quantization ,chauhan2018performance}} \\ \cline{3-3} 
& \multicolumn{1}{l|}{} & \multicolumn{1}{l|}{Multiplexing ~\cite{han2016eie, eshratifar2020run, zhang2019mbfn}}                                                              \\ \cline{3-3} 
& \multicolumn{1}{l|}{}& \multicolumn{1}{l|}{Weight sharing ~\cite{han2015deep,han2016eie,he2018multi,wu2018sharing,georgiev2017low}}                                                                   \\ \cline{2-3} 
& \multicolumn{2}{l}{}                                                                                                                                                                 \\ \cline{2-3} 

& \multicolumn{1}{l|}{\multirow{3}{*}{Bits precision}}
& \multicolumn{1}{l|}{Estimation using integer ~\cite{jacob2018quantization,lee2019neuro}} \\ \cline{3-3} 
& \multicolumn{1}{l|}{} & \multicolumn{1}{l|}{Low bits representation ~\cite{hubara2017quantized}}                                                              \\ \cline{3-3} 
& \multicolumn{1}{l|}{}& \multicolumn{1}{l|}{Binarization ~\cite{hubara2016binarized,yang2019cdeeparch,umuroglu2017finn,louizos2017bayesian}}                                                                   \\ \cline{2-3} 
& \multicolumn{2}{l}{}                                                                                                                                                                 \\ \cline{2-3} 

& \multicolumn{1}{l|}{\multirow{3}{*}{Knowledge distillation}}
& \multicolumn{1}{l|}{Logits transfer ~\cite{wang2019private,hinton2015distilling,yun2020regularizing,li2020few,cheng2020explaining}} \\ \cline{3-3} 
& \multicolumn{1}{l|}{} & \multicolumn{1}{l|}{Teaching assistant ~\cite{mirzadeh2019improved}}                                                              \\ \cline{3-3} 
& \multicolumn{1}{l|}{}& \multicolumn{1}{l|}{Domain adaptation~\cite{wang2019private,duong2019shrinkteanet,yang2020mobileda}}                                                                   \\ \cline{2-3} 
& \multicolumn{2}{l}{}                                                                                                                                                                 \\ \cline{2-3} 

& \multicolumn{1}{l|}{\multirow{1}{*}{Miscellaneous}}
& \multicolumn{1}{l|}{\cite{yao2017deepsense, yao2018fastdeepiot,abc,radu2016towards,fang2018nestdnn,radu2018multimodal,mathur2017deepeye,xu2018deepcache,shen2019deepapp,xue2019deepfusion,svyatkovskiy2017training,wang2018deep,zhang2018shufflenet}}  
  \\ \hline

\end{tabular}
\end{table*}

\section{Network Pruning}\label{np}
\begin{figure}[h]
 \centering
 \includegraphics[scale=1.0]{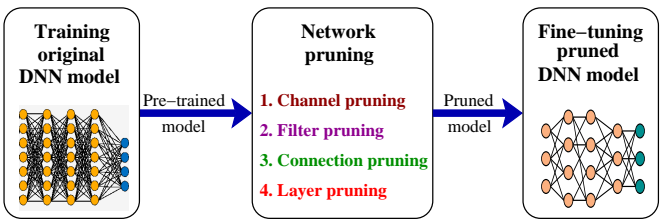}
 \caption{Categories and steps involved in network pruning.}
 \label{net_pru}
\end{figure}

This section discusses the network pruning techniques for DNN compression. The key idea is to remove unimportant components such as layers, filters, channels, \textit{etc.}, from the original DNN model. The remaining components form a compressed DNN model. The compressed model requires both minimal processing resources and consume lower storage than the original DNN model. Further, the compressed model is trained on the existing dataset for a colossal number of epochs that leads to the fine-tuning of the model. The two major research challenges associated with the network pruning are mitigating the accuracy compromise due to network compression and reducing the time in fine-tuning of the DNN model~\cite{NIPS2017_6910,Mallya}. We divide the network pruning into four categories depending upon the existing work, \textit{i.e.,} Channel pruning~\cite{he2017channel,liu2018improvement,chen2019evolutionary,liu2017learning, yan2019vargfacenet,howard2017mobilenets}, Filter pruning~\cite{luo2017thinet, bhattacharya2016sparsification, zhou2019knee, huynh2017deepmon,denton2014exploiting}, Connection pruning~\cite{han2015deep,yao2017deepiot,han2016eie,he2018multi,chen2019evolutionary,qin2018compress, parashar2017scnn,molchanov2017variational,louizos2017bayesian,guo2016dynamic}, and Layer pruning~\cite{he2018multi,tan2019mnasnet, chauhan2018performance}. Fig.~\ref{net_pru} illustrate the categories and step involved in the network pruning. Table~\ref{table2} summarizes the existing literature on network pruning under different categories. The detailed description of different network pruning techniques are as follows.

\begin{table*}[h]
\caption{Summary of network pruning techniques for DNN compression.}
\resizebox{1.0\textwidth}{!}{
\begin{tabular}{|c|c|l|l|l|l|l|l|}
\hline
\multirow{2}{*}{\textbf{Paper}}                              & \multirow{2}{*}{\textbf{\begin{tabular}[c]{@{}c@{}}Abbreviated \\ name\end{tabular}}}  & \multicolumn{1}{c|}{\multirow{2}{*}{\textbf{Address the challenge}}}                                                      & \multicolumn{1}{c|}{\multirow{2}{*}{\textbf{Proposed solution}}}                                                         & \multicolumn{3}{c|}{\textbf{Suitable for}}                            & \multirow{2}{*}{\textbf{Category}}                                                                                                                                                     \\ \cline{5-7} 
                                                             &                                            & \multicolumn{1}{c|}{}                                                                                                     & \multicolumn{1}{c|}{}                                                                                                    & \textbf{CNN}          & \textbf{FC}           & \textbf{RNN}          &                                                              \\ \hline
                                                             
%%%%%%%%%%%%%%%%%%%%%%%%%%%%%%%%                                                             
~\cite{he2017channel}                  & \textemdash                 & To accelerate deep CNN for faster inference                                                                               &  Efficiently removing redundant channels                                   & \cmark & \xmark & \xmark & \multicolumn{1}{c|}{\multirow{10}{*}{\begin{tabular}[c]{@{}c@{}}Channel\\ pruning\end{tabular}}}                            \\ \cline{1-7}

~\cite{liu2018improvement}             & \textemdash                 & Improving channel pruning scheme                                                                                          & \begin{tabular}[c]{@{}l@{}}Determine pruning threshold prior to\\ actual pruning\end{tabular}  & \cmark & \xmark & \xmark &                                                                                                                                                                                  \\ \cline{1-7}

~\cite{chen2019evolutionary}           & EvoNAS                      & \begin{tabular}[c]{@{}l@{}}Resolve dependency between channels and \\ neurons\end{tabular}                                & \begin{tabular}[c]{@{}l@{}}Combinatorial optimization for network \\ pruning\end{tabular}                                & \cmark & \cmark & \xmark &                                                                                                                                                                                  \\ \cline{1-7}

~\cite{liu2017learning}                & \textemdash                 & \begin{tabular}[c]{@{}l@{}}To decrease model-size, runtime, memory, \\ and computation\end{tabular}                    & \begin{tabular}[c]{@{}l@{}}Network slimming using channel \\ level sparsity\end{tabular}                                 & \cmark & \xmark & \xmark &                                                                                                                                                                                   \\ \cline{1-7}

~\cite{yan2019vargfacenet}             & VarGNet                                    & To reduce the operation in CNN                                                                                            & \begin{tabular}[c]{@{}l@{}}Variable group convolution and angular \\ distillation loss\end{tabular}                      & \cmark & \xmark & \xmark &                                                                                                                                                                                   \\ \cline{1-7}

~\cite{howard2017mobilenets}           & MobileNet                                  & To build a light-weight DNN model                                                                                         & \begin{tabular}[c]{@{}l@{}}Deep-wise and $1\times1$ convolutions\end{tabular}                              & \cmark & \xmark & \xmark &                                                                                                                                                                                  \\ \hline
%%%%%%%%%%%%%%%%%%%%%%%%%%%%%%%%%%%%%%%%%%%%%%%
~\cite{luo2017thinet}                  & ThiNet                                     & \begin{tabular}[c]{@{}l@{}}Efficient framework for accelerating operations \\ in CNN\end{tabular}                      & \begin{tabular}[c]{@{}l@{}}Discarding unimportant filters using \\ statistical information\end{tabular}                  & \cmark & \xmark & \xmark & \multicolumn{1}{c|}{\multirow{8}{*}{\begin{tabular}[c]{@{}c@{}}Filter\\ pruning\end{tabular}}}                                                                                                        \\ \cline{1-7}

~\cite{bhattacharya2016sparsification} & \textemdash                 & \begin{tabular}[c]{@{}l@{}}To reduce the storage and memory requirement \\ during training and inference\end{tabular}      & \begin{tabular}[c]{@{}l@{}}Sparsification in fully connected layer \\ and convolutional filters\end{tabular}             & \cmark & \cmark & \xmark &                                                                                                                                                                                   \\ \cline{1-7}

~\cite{zhou2019knee}                   & \textemdash                 & \begin{tabular}[c]{@{}l@{}}Prune parameters that results in minimal\\ performance degradation\end{tabular}                & \begin{tabular}[c]{@{}l@{}}Knee-guided evolutionary algorithm\\ for optimal filter pruning.\end{tabular}                 & \cmark & \xmark & \xmark &                                                                                                                                                                                  \\ \cline{1-7}

~\cite{huynh2017deepmon}                     & DeepMon                                    & \begin{tabular}[c]{@{}l@{}}To provide deep learning inference on mobile \\ devices\end{tabular}                           & \begin{tabular}[c]{@{}l@{}}An optimization mechanism for \\ convolutional operation on mobile\end{tabular}               & \cmark & \xmark & \xmark &                                                                                                                                                                                   \\ \cline{1-7}

~\cite{denton2014exploiting}           & \textemdash                 & \begin{tabular}[c]{@{}l@{}}To speed up the test-time evaluation of the \\ large DNN model\end{tabular}                 & \begin{tabular}[c]{@{}l@{}}Finds appropriate low-rank approximation \\ for convolutional filters\end{tabular}         & \cmark & \xmark & \xmark &                                                                                                                                                                                   \\ \hline

%%%%%%%%%%%%%%%%%%%%%%%%%%%%%%%%%%%%%%%%%%%%%%%%%%%%%%%%%%%%%%%%%%%%%

~\cite{han2015deep}                    & \textemdash                 & \begin{tabular}[c]{@{}l@{}}To reduce storage and bandwidth of  DNN \\ model with minimal accuracy compromise\end{tabular} & \begin{tabular}[c]{@{}l@{}}Quantization using clustering \\ technique, and further Huffman coding\end{tabular}           & \cmark & \cmark & \cmark & \multicolumn{1}{c|}{\multirow{16}{*}{\begin{tabular}[c]{@{}c@{}}Connection \\ pruning\end{tabular}}}                                                                             \\ \cline{1-7}

~\cite{yao2017deepiot}                 & DeepIoT                                    & \begin{tabular}[c]{@{}l@{}}To ensure the deployment of deep learning \\ models on IoT devices\end{tabular}                & Optimal dropout estimation                                                                                               & \cmark & \cmark & \cmark & \multicolumn{1}{c|}{}                                                                                                                                                            \\ \cline{1-7}

~\cite{han2016eie}                     & EIE                                        & \begin{tabular}[c]{@{}l@{}}To perform DNN based inference on the \\ compressed network\end{tabular}                       & \begin{tabular}[c]{@{}l@{}}Pruning of redundant connections  and \\ weight sharing among different elements\end{tabular} & \cmark & \cmark & \cmark & \multicolumn{1}{c|}{}                                                                                                                                                             \\ \cline{1-7}

~\cite{he2018multi}                    & MTZ                                        & \begin{tabular}[c]{@{}l@{}}To induced a minimal change in error upon \\ layer sharing\end{tabular}                        & \begin{tabular}[c]{@{}l@{}}Layer-wise neuron sharing and subsequent \\ weight update\end{tabular}                        & \cmark & \cmark & \cmark & \multicolumn{1}{c|}{}                                                                                                                                                             \\ \cline{1-7}

~\cite{qin2018compress}                & \textemdash                 & \begin{tabular}[c]{@{}l@{}}To provides a better selection mechanism \\ for deployment of DNN model\end{tabular}           & Guided deployment for wiser pruning                                                                &                       \cmark & \cmark & \cmark & \multicolumn{1}{c|}{}                                                                                                                                                             \\ \cline{1-7}

~\cite{parashar2017scnn}               & SCNN                                       & \begin{tabular}[c]{@{}l@{}}To improve performance and provides \\ energy efficiency\end{tabular}                          & \begin{tabular}[c]{@{}l@{}}Zero-valued weight estimation generated \\ from ReLU operation\end{tabular}                   & \cmark & \xmark & \xmark & \multicolumn{1}{c|}{}                                                                                                                                                            \\ \cline{1-7}

~\cite{molchanov2017variational}       & \textemdash                 & \begin{tabular}[c]{@{}l@{}}A sparse solution for compressing DNN models\end{tabular}                               & Using  sparse variational dropout                                                                                        & \cmark & \xmark & \xmark & \multicolumn{1}{c|}{}                                                                                                                                                             \\ \cline{1-7}

~\cite{louizos2017bayesian}            & \textemdash                 & \begin{tabular}[c]{@{}l@{}}Hierarchical priority to prune the nodes inside\\  the DNN model\end{tabular}                  & Bayesian compression for Deep learning                                                                                   & \cmark & \cmark & \cmark & \multicolumn{1}{c|}{}                                                                                                                                                             \\ \cline{1-7}

~\cite{han2015learning}                & \textemdash                 & \begin{tabular}[c]{@{}l@{}}Pruning method for DNN model by removing \\ unimportant connections\end{tabular}               & Using connection pruning threshold                                                                                       & \cmark & \cmark & \cmark & \multicolumn{1}{c|}{}                                                                                                                                                             \\ \cline{1-6}

~\cite{guo2016dynamic}                 & \textemdash                 & \begin{tabular}[c]{@{}l@{}}Solving irretrievable network damage due to \\ incorrect pruning\end{tabular}                  & \begin{tabular}[c]{@{}l@{}}Splicing into the network compression to \\ avoid incorrect pruning\end{tabular}              & \cmark & \xmark & \xmark & \multicolumn{1}{c|}{}                                                                                                                                                            \\ \hline
%%%%%%%%%%%%%%%%%%%%%%%%%%%%%%%%%%%%%%%%%%%%%%%%%%%%%%%%%%%%%%%%%%%%%%%%%%%%%%%%%%%%%%%%%%%%%%%%
~\cite{tan2019mnasnet}                & FINN                       & \begin{tabular}[c]{@{}l@{}}To build a fast and flexible heterogeneous \\ DNN architecture\end{tabular}                    & \begin{tabular}[c]{@{}l@{}}Using a set of optimization for mapping \\ binarized neural networks to hardware\end{tabular} & \cmark & \cmark & \xmark & \multicolumn{1}{c|}{\multirow{5}{*}{\begin{tabular}[c]{@{}c@{}}Layer\\ pruning\end{tabular}}}                                                                                    \\ \cline{1-7}

\cite{chauhan2018performance}         & \textemdash                  & \begin{tabular}[c]{@{}l@{}}Pruning method for deploying DNN model\\ on embedded devices\end{tabular}                      & \begin{tabular}[c]{@{}l@{}}Singular value decomposition based \\ factorization method\end{tabular}                       & \xmark & \xmark & \cmark & \multicolumn{1}{c|}{}                                                                                                                                                            \\ \cline{1-7}

~\cite{he2018multi}                    & MTZ                          & \begin{tabular}[c]{@{}l@{}}Minimal change in error upon layer sharing\end{tabular}                        & \begin{tabular}[c]{@{}l@{}}Layer pruning and weight update\end{tabular}                        & \cmark & \cmark & \cmark & \multicolumn{1}{c|}{}                                                                                                                                                            \\ \hline
%%%%%%%%%%%%%%%%%%%%%%%%%%%%%%%%%%%%%%%%%%%%%%%%%%%%%%%%%%%%%%%%%%%%%%%%%%%%%%%%%%%%%%%%%%%%%%%%%%%%%%%%%%%%%%%%%%%%%
\end{tabular}
}
\label{table2}
\end{table*}

\subsection{Channel pruning}\label{cp}
In this section, we provide an overview of different channel pruning techniques discussed in the existing literature. Channel pruning is a concept of reducing the number of channels in the input supplied to the intermediate layers of the DNN model~\cite{peng2019collaborative}. The data supplied to the DNN model are initially channelized to form an appropriate input, \textit{e.g.} image have $3$ channels (RGB). The output of each layer of the DNN model also contains different channels that improve the model performance but increases the computation and storage. Therefore, it is beneficial to remove unimportant channels to reduce the computation and storage requirements. Authors in~\cite{he2017channel} proposed an inference time approach to perform channel pruning by effectively removing the redundant channels from CNN. They claim to accelerate the deep CNN using two sequential steps, including, regression-based channel selection followed by estimation of least reconstruction error. Authors have proved their proposed mechanism for accelerating deep CNN operations using compact architecture by performing experiments on different datasets. The reconstruction error in this approach is formulated as follows. 

Let $k$ represent the number of channels in a feature map which we have to pure to reduce the number of channels to $k{'}$. Let a convolutional filter $\mathbf{F}$ of size $o\times k \times f_h \times f_w$ applied to a input $\mathbf{I}$ of dimension $N\times k \times f_h \times f_w$ to produce output matrix $\mathbf{O}$. Here, $N$ denotes the number of samples in dataset $\mathcal{D}$ and  $f_h\times f_w$ is the dimension of the convolutional kernel. The minimization problem is formed as
\begin{align}\nonumber
 \arg \min_{\delta, \mathbf{F}} \frac{1}{N}\Big\| \mathbf{O} - \sum_{j=1}^{k}\delta_j \mathbf{F}_j\mathbf{I}_j\Big\|_F^2,\\
 \text{subject to, } ||\delta||_0 \leq k{'},
\end{align}
where, $||\cdot||_F$ denotes the Frobenius norm. $\mathbf{I}_j$ represents the data input for channel $j,1\leq j \leq k$ and $\mathbf{F}_j$ is the $j^{th}$ slice of $\mathbf{F}$. $\delta$ a coefficient vector of length $k$ used in channel selection, $\delta_j$ is the $j^{th}$ entry of $\delta$. Liu \textit{et.al.}~\cite{liu2018improvement} proposed an improved scheme of pruning operation, where, at the initial step, the pruning threshold is determined by analyzing the network connections. Further, those connections and channels are removed that are less than the pruning threshold. The threshold estimation in the proposed scheme preserves the accuracy of the pruned network up to some extent. In~\cite{chen2019evolutionary} authors claim that the prior assumption of channels and neurons independent in the deep neural network is wrong. They claim that there are certain dependencies among channel and neurons. Authors formed a combinatorial optimization problem for network pruning which is solved using an evolutionary algorithm and attention mechanism. They named the proposed mechanism as Evolutionary NetArchitecture Search (EvoNAS). Authors in~\cite{liu2017learning} proposed a novel learning scheme for CNN that decrease model-size, runtime memory, and computing operations. They claim to retain the accuracy of the compressed model. The proposed scheme enforces channel-level sparsity in the network and dierctly applies to CNN architectures. Therefore, introduces minimal  overhead during the training of the deep learning model. They named their approach as network slimming that compressed large model to a compact model without a higher accuracy compromise. The main idea behind this approach is to introduce a scaling factor ($\Delta$) for each channel in the deep learning model. Next, the model and $\Delta$ are jointly trained using the sparsity regularization.The training objective is defined as

\begin{equation}
 Loss = \sum_{\mathbf{x},y} \mathcal{L}(f(\mathbf{x},\mathbf{W}),y) + \mu \sum_{\Delta} g(\Delta),
\end{equation}
where, $(\mathbf{x}, y)$ denotes the training input and output, $\mathbf{W}$ represents the weight matrix, and $\mathcal{L}(\cdot)$ is the training loss imposed by the model. $g(\cdot)$ denotes the penalty function and $\mu$ balances the difference between two terms. 

Yan \textit{et. al.}~\cite{yan2019vargfacenet} proposed an approach (named VarGNet) for reducing the operations in the CNN model that uses variable group convolution. The use of variable group convolution helps in solving the conflict between small computational cost and the unbalanced computational intensity. Further, they adopt angular distillation loss to improve the performance of the generated lightweight model. Authors in~\cite{howard2017mobilenets} proposed a DNN compression technique based on the streamlined architecture that uses depth-wise separable convolutions to build a lightweight model. The authors named the technique as MobileNets that estimates two hyperparameters, \textit{i.e.,} width multiplier and resolution multiplier.  It allows the model developer to choose a small network that matches the resources restrictions (latency, size) for different applications. In MobileNets, the depthwise convolution applies a single filter to each input channels. Further, $1 \times 1$ convolution is performed to combine inputs into a new set of output. MobileNets achieve faster inference by exploiting the sparsity of the dataset.  The role of the width multiplier $wd$ is to thin a network uniformly at each layer. For a given layer, the number of input channels $M$ becomes $(wd)M$. The resolution multiplier $rm$ is applied to the input image and the same multiplier subsequently reduces the internal representation of every layer.

\noindent $\bullet$ \textbf{Remarks:} The existing literature on channel pruning, EvoNAS~\cite{chen2019evolutionary} VarGNet~\cite{yan2019vargfacenet}, MobileNets~\cite{howard2017mobilenets},\cite{he2017channel,liu2018improvement,liu2017learning} have covered different mechanisms of channel pruning. However, none of the existing work helps in diminishing the accuracy compromise due to channel pruning. Apart from the accuracy compromise, the existing literature misses the effect of a channel elimination on the performance of different layers of the DNN model. As, removal of single channel in one layer can also effect other layers in the DNN model.

\subsection{Filter pruning}\label{fp}
\begin{figure}[h]
 \centering
 \includegraphics[scale=1.0]{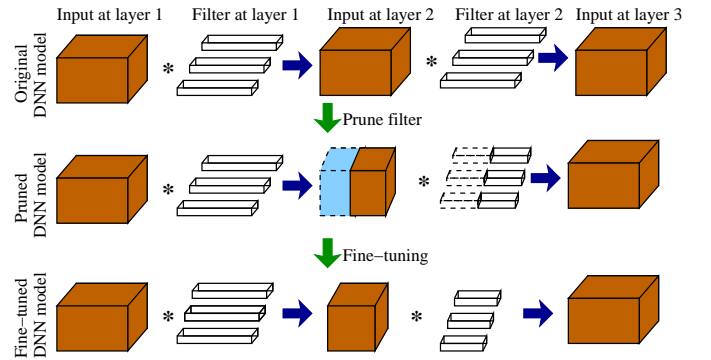}
 \caption{An example scenario of the filter level pruning~\cite{luo2017thinet}.}
 \label{filter_c}
\end{figure}

The convolutional operation in the CNN model incorporates a large number of filters to improve its performance under different processes of classifications, regressions, and predictions. As it assumed that the increment in the number of filters, improves the distinguishable characteristics of the spatial features generated from the CNN model.~\cite{singh2019stability}. However, this increment in the convolutional filters leads to a significant increase in the number of floating-point operations in the DNN model. Therefore, the elimination of the unimportant filters plays a decisive role in reducing the computational requirements of the DNN model. An example scenario of filter level pruning is illustrated in Fig.~\ref{filter_c}. The existing work on filter pruning~\cite{luo2017thinet, bhattacharya2016sparsification, zhou2019knee, huynh2017deepmon,denton2014exploiting} are discussed as follows.

Authors in~\cite{luo2017thinet} proposed an efficient framework, namely ThiNet, for accelerating the operation of the CNN model using compression in both training and testing phases. They have incorporated filter-level pruning, where, an unimportant filter is discarded based on the statistical information computed form the next layer. Authors have established a filter level pruning as an optimization problem to determine the filters that are to be pruned. ThiNet uses a greedy algorithm for solving the optimization problem that is defined as follows 

\begin{align}
 \arg \min_{E} \sum_{i=1}^{N}\Big(\hat{y}_i-\sum_{j\in E}\hat{\mathbf{x}}_{ij}\Big)^2,\\ 
 \text{subject to, } |E|=k\times c_{rate},\\
 E \subset \{1,2,\cdots,k\},
\end{align}
where, $N$ is the number of training examples ($\hat{\mathbf{x}}_i,\hat{y}_i$), $|E|$ is the number of elements in a subset of E, $k$ denotes number of channel in CNN model, and $c_{rate}$ is compression rate that determines number of channels retrained after compression.

Bhattacharya \textit{et. al.}~\cite{bhattacharya2016sparsification} proposed a DNN compression technique that removes the sparsification in fully connected layer and convolutional filters. The main motive was to reduce the storage requirement of device and processor resources in both training and inference phases. The basic operational principle of this work is the hypothesis that computational and space complexity of DNN model can be considerably increase using sparsification of layers and convolutional filters.

In~\cite{zhou2019knee} authors, defined the filter pruning as a multi-objective optimization problem followed by a knee-guided evolutionary algorithm for its solution. They established a trade-off between the number of parameters and performance compromise due to model compression. The basic principle of this mechanism is to prune parameters that result in little performance degradation. To estimate the importance of a parameter, they have used the criteria of performance loss to identify the redundancy. To achieve a compressed model with a small size, the number of filters preserved must be less as possible to achieve a considerable accuracy. The problem can be treated as to find a compact binary representation that can prune maximum filters while preserving the performance to a greater extent. This pruning mechanism falls under the group of filter pruning in the CNN, as it simultaneously poses the advantage of parameter reduction and minimizing computational overhead. This mechanism can be further improved by incorporating the low-rank estimation. Authors in~\cite{huynh2017deepmon} proposed a mechanism named DeepMon to provide deep learning inference on mobile devices. They claim to run the inference in limited time and provides energy efficiency using the Graphical Processing Unit (GPU) on the mobile device. The authors proposed an optimization mechanism for processing convolutional operation on mobile GPU. The mechanism utilizes the internal processing structure of CNN incorporating filters and the number of connections to reuse the results. Thus, removing unimportant filters and connections for proving faster inference.

Authors in~\cite{denton2014exploiting} speed up the test-time evaluation of the large DNN model, designed for object recognition tasks. The authors have exploited the redundancy present within the convolutional filters to derive approximations that significantly reduce the required computation. They have started compression of each convolutional layer by finding an appropriate low-rank approximation, and then fine-tune until the prediction performance is restored.

\noindent $\bullet$ \textbf{Remarks:} The filter pruning technique presented in the existing literature~\cite{luo2017thinet, bhattacharya2016sparsification, zhou2019knee, huynh2017deepmon,denton2014exploiting} have successfully reduced the number of floating-point operations in the CNN. However, the fully connected layers and recurrent layers also contribute a major portion of floating-point operations in the DNN model, which should not be ignored during DNN compression. Additionally, the existing work cannot be employed on the DNN model that is having a large number of layers with minimal filters on each convolutional layer.

\subsection{Connection pruning}
\begin{figure}[h]
 \centering
 \includegraphics[scale=1.00]{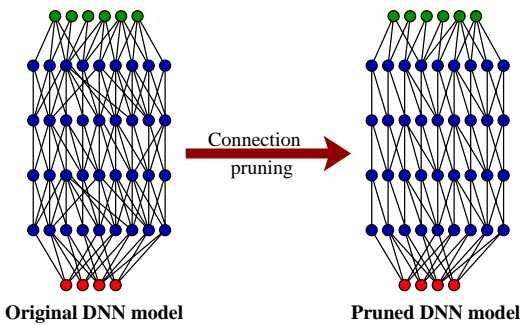}
  \caption{An illustration of connection pruning in the DNN model.}
 \label{connection_p}
\end{figure}

The number of input and output connections to a layer of DNN model determines the number of parameters. These parameters can be used to estimate the storage and computation requirement of the DNN model~\cite{9130098}. Since the DNN model requires a large number of parameters in their operations; therefore, it is convenient to reduce the parameters by eliminating unimportant connection for different layers in the DNN model~\cite{siebel2009efficient}. The existing studies on the connection pruning~\cite{han2015deep,yao2017deepiot,han2016eie,he2018multi,chen2019evolutionary,qin2018compress,  parashar2017scnn,molchanov2017variational,louizos2017bayesian,guo2016dynamic} have attempted to remove the unimportant connection from the DNN models. Fig.~\ref{connection_p} illustrates an example scenario of connection pruning in the DNN model. The overview of different connection pruning techniques discussed in the  existing literature are as follows

Authors in~\cite{han2015deep} proposed a connection pruning technique that first prunes the DNN model by learning only important connection using dropout. Next, the learned weights are quantized using clustering technique, and further Huffman coding is applied to reduce the storage requirement. The authors under this technique try to fit the model into cache memory rather than main memory. In this way, they encourage us to run applications on tiny computing devices with limited storage and energy. Here, the goal is to reduce the storage and bandwidth of the DNN model with minimal accuracy compromise. It also justified the goal of tiny device execution, such as better privacy, uses least network bandwidth and provides output in real-time. 

In~\cite{yao2017deepiot} authors proposed an approach that helps in the deployment of deep learning models on IoT devices. Thus, named the approach as DeepIoT. The proposed approach is capable of performing compression on commonly used deep learning structures, including CNN, fully connected layers, and recurrent neural network. DeepIoT incorporates dropout techniques to remove unimportant connections from the network. Here, the chosen dropout is not a fixed value but optimally determined for generating optimal compressed networks. In DeepIoT, small dense matrices are obtained by performing compression of the different structures in the network. These dense matrices contain a minimal number of non-redundant elements, such as input and output dimension of a layer, filters, \textit{etc}. The process of obtaining dense matrices must safeguard the network from higher accuracy compromise.

Han \textit{et. al.}~\cite{han2016eie} proposed an Energy efficient Inference Engine (EIE) to perform DNN based inference on the compressed network obtained after pruning of redundant connections and weight sharing among different elements. The EIE accelerates the sparse matrix multiplication by adopting weight sharing without losing the accuracy of the model. In EIE, the processing is performed using an array of processing elements. Each element simultaneously partitions the network in SRAM and perform the computation of their respective part. In~\cite{he2018multi} authors proposed a Multi-Tasking Zipping (MTZ) framework that automatically merges multiple correlated DNN models to perform cross model compression. In MTZ, the compression is performed using layer-wise neuron sharing and subsequent weight update. Here, the authors induced a minimal change in error upon layer sharing. 

Authors in~\cite{qin2018compress} developed a quantitative characterization approach for a deep learning model to make its deployment on embedded devices. It not only provides a better selection mechanism for deployment of DNN model on the embedded device but also guides the development of compression technique using a wiser connection pruning. In~\cite{parashar2017scnn} authors proposed a zero-valued weight estimation from a network during training that is generated from ReLU operation. The authors named the approach as Sparse CNN (SCNN). It is a DNN compression architecture that improves performance and provides energy efficiency. SCNN uses both weight and activation sparsity, which enhances the power of the compressed DNN. SCNN also employ an effective dataflow mechanism inside the DNN that maintains the sparse weights and activations in the DNN compression. It further reduces unimportant data transmission and storage requirement.

Authors in~\cite{molchanov2017variational} proposed a sparse variational dropout, which is an extension of variational dropout. Here, all possible values of dropouts are considered, which leads to a sparse solution for compressing the DNN models. Authors have provided an approximation technique that incorporates KL-divergence in variational dropout. They experimentally demonstrate that higher sparsity is achieved in CNN and fully connected layers.  Let $N$ denotes the number of instances in dataset $\mathcal{D}$, whose $i^{th}$ instance is denoted as $(\mathbf{x}_i,y_i)_{i=1}^N$. The goal of a learning mechanism is to estimate parameter $\theta$ of a model $p(y/\mathbf{x},\theta)$. When we incorporate Bayesian learning after data $\mathcal{D}$ arrival, the prior distribution is transformed into a posterior distribution, $p(\theta/\mathcal{D})=p(\mathcal{D/\theta})p(\theta)/p(\mathcal{D})$. In variational inference the posterior distribution $p(\theta/\mathcal{D})$ is approximation using distribution $r_{\phi}(\theta)$. The authors uses KL divergence $D_{KL}(r_{\phi}(\theta)||p(\theta/\mathcal{D}))$ to estimate quality of approximation. The optimal value of parameter $\phi$ is obtained as follows

\begin{align}\nonumber
 \mathcal{L}(\phi) = \mathcal{L}_D(\phi) - D_{KL}(r_{\phi}(\theta)||p(\theta)),\\
\text{where, } \mathcal{L}_D(\phi) = \sum_{i=1}^N \mathbb{E}_{r_{\phi}(\theta)}[\log p(y_i/\mathbf{x}_i,\theta)].
\end{align}

Louizos \textit{et. al.}~\cite{louizos2017bayesian} proposed a Bayesian compression for Deep learning. They introduced a hierarchical priority to prune the nodes inside the DNN model inspite weight reduction. The authors have incorporated posterior uncertainties for determining the fixed point precision to encode the weights. Further, the authors employ inducing sparsity for hidden neuron despite individual weights to prune neurons. In~\cite{han2015learning} authors proposed a network pruning method for DNN model that removes unimportant connections. The proposed pruning method first identifies the unimportant connections having weight less than a given threshold. Next, these connections are removed from the DNN model. Further, fine-tuning is performed with the remaining connections.

Authors in~\cite{guo2016dynamic} proposed a network compression technique to perform network pruning. The pruning incorporates deleting unimportant parameters and retaining the remaining ones. The authors have incorporated splicing into the network compression to avoid incorrect pruning.  There are two major issues related to network interconnections in the DNN model. The First issue is irretrievable network damage due to incorrect pruning, and second is the inconsistency of the DNN model due to an inefficient machine. The authors have claimed to solve both shortcomings of the network pruning through continuous pruning and splicing.

\noindent $\bullet$ \textbf{Remarks:} The connection pruning technique provides a high order compression of the DNN model in contrast with channel pruning and filter pruning discussed in Section~\ref{cp} and \ref{fp}, respectively. The existing studies, DeepIoT~\cite{yao2017deepiot}, EIE~\cite{han2016eie}, MTZ~\cite{he2018multi}, SCNN~\cite{parashar2017scnn}, \cite{han2015deep,chen2019evolutionary,qin2018compress, molchanov2017variational,louizos2017bayesian,guo2016dynamic} have emphasized on connection pruning to perform DNN compression. However, a major concern of connection pruning technique is the elimination of important connection during DNN compression, needs more emphasis, such as pruning and splicing discussed in~\cite{guo2016dynamic}. Additionally, the connection pruning suffers from higher accuracy compromise when the network is dynamically building and rebuilding on RCDs.

\subsection{Layer pruning}
The last category of the network pruning techniques is layer pruning, where, some selected layers from the network are removed to compress the DNN model. The layer pruning is highly utilized for deploying DNN model on tiny computing devices, where, we need an ultra-high compression of DNN model~\cite{liu2020autocompress}. The primary issue with layer pruning is the loss of the semantic structure of the DNN model that generates low-quality features. These low-quality features result in performance inefficiency. The prior studies on layer pruning~\cite{he2018multi,tan2019mnasnet, chauhan2018performance} for DNN compression are elaborated as follows.

Authors in~\cite{tan2019mnasnet} presented a framework that builds fast and flexible heterogeneous architecture, named FINN. They utilized a set of optimization for mapping binarized neural networks to the hardware. The authors have performed the implementation of different DNN components, including, convolutional, pooling, and fully connected layers. The implementation was conducted in such a manner that meet out the performance demand of the users. They claim to perform millions of classifications per second with microsecond latency on embedded devices.In~\cite{chauhan2018performance} authors compared the performance of different classification approaches and proposed a pruning method for deploying DNN model on embedded devices. They have taken three embedded devices, including smartphone, smartwatch, and Raspberry Pi, for deploying compressed LSTM model. They claim to achieve significant accuracy by performing compression up to $25\%$.

The authors have used the Singular Value Decomposition (SVD) based factorization method that allows the decomposition of weight matrix ($\mathbf{W}$) into three sub-matrices, \textit{i.e.,} $\mathbf{A}$, $\mathbf{E}$, and $\mathbf{B}$ of dimensions  $N\times l$, $l\times l$, and $l\times M$, respectively. Here, $N$ denotes the number of instance in dataset $\mathcal{D}$ having sample length of size $M$. $l$ denotes the number of class labels in the dataset and matrix $\mathbf{E}$ is a diagonal matrix. The decomposition of the weight matrix is represented as 
\begin{align}
 \mathbf{W}_{N\times M} = \mathbf{A}_{N \times l} \mathbf{E}_{l \times l} \mathbf{B}_{l\times M}.
\end{align}

The weight factorization helps in achieving both storage gain $\mathscr{S}_g$ and computation gain $\mathscr{C}_g$ defined as follows

\begin{align}
 \mathscr{S}_g = \frac{N \times M}{N\times l + l^2 + l \times M}.\\
 \mathbf{C}_g = \frac{N^2 \times M}{N^2\times l + l^3 + l^2\times M}.
\end{align}

\noindent $\bullet$ \textbf{Remarks:} The existing literature on layer pruning, FINN~\cite{tan2019mnasnet}, \cite{he2018multi,chauhan2018performance} have reduced both storage and computation requirement of DNN model by providing an ultra high-level pruning. However, the layer pruning results in higher accuracy compromise due to structural deterioration of the DNN model, which should be mitigated to enhance the utility of layer pruning.

\section{Sparse representation}\label{sr}
In the previous section, we have covered those techniques for DNN compression that prunes unimportant components, including, layers, filters, and channels, \textit{etc}. This section, on the other hand, gives an overview of the DNN compression techniques that preserve the overall structure of the DNN model. Here, the sparsity in the representation of the DNN model is exploited to reduce both storage and processing requirement of the DNN model~\cite{zhang2015survey, zhang2011kernel}. The sparsity in the DNN model persists in the weight matrices due to the following two reasons

\begin{enumerate}
 \item The value of stored weights is zero or near to zero. It could be beneficial to remove these weights to reduce the computation and storage requirements.
 \item The value of maximum stored weights are alike that provides a convenience to replace multiple weights having single connections with single weight having multiplex connections.
\end{enumerate}

The existing literature on sparse representation~\cite{han2015deep,qin2018compress,jacob2018quantization ,chauhan2018performance, han2016eie, eshratifar2020run, zhang2019mbfn, han2015deep,han2016eie,he2018multi,wu2018sharing,georgiev2017low} have incorporated above two reasons for compressing the DNN models. We categorize the existing literature in three parts, \textit{i.e.,} Quantization ~\cite{han2015deep,qin2018compress,jacob2018quantization ,chauhan2018performance}, Multiplexing ~\cite{han2016eie, eshratifar2020run, zhang2019mbfn}, and Weight sharing ~\cite{han2015deep,han2016eie,he2018multi,wu2018sharing,georgiev2017low}, as illustrated in Table~\ref{table3}. The overview of existing literature on these categories are as follows.

\begin{table*}[h]
\caption{Illustration of DNN compression techniques that incorporates sparse representation.}
\resizebox{1.0\textwidth}{!}{
\begin{tabular}{|c|c|l|l|l|l|l|l|}
\hline
\multirow{2}{*}{\textbf{Paper}}                              & \multirow{2}{*}{\textbf{\begin{tabular}[c]{@{}c@{}}Abbreviated \\ name\end{tabular}}}  & \multicolumn{1}{c|}{\multirow{2}{*}{\textbf{Address the challenge}}}                                                      & \multicolumn{1}{c|}{\multirow{2}{*}{\textbf{Proposed solution}}}                                                         & \multicolumn{3}{c|}{\textbf{Suitable for}}                            & \multirow{2}{*}{\textbf{Category}}                                                                                                                                                     \\ \cline{5-7} 
                                                             &                                            & \multicolumn{1}{c|}{}                                                                                                     & \multicolumn{1}{c|}{}                                                                                                    & \textbf{CNN}          & \textbf{FC}           & \textbf{RNN}          &                                                              \\ \hline
                                                             
%%%%%%%%%%%%%%%%%%%%%%%%%%%%%%%%                                                             
~\cite{han2015deep}           & \textemdash                           & \begin{tabular}[c]{@{}l@{}}To compress DNN by reducing the bits\\  required for weight matrices representation\end{tabular}           & \begin{tabular}[c]{@{}l@{}}Weight quantization to reduce the \\ storage requirement of DNN model\end{tabular}                                & \cmark & \cmark & \cmark & \multicolumn{1}{c|}{\multirow{7}{*}{Quantization}}                                              \\ \cline{1-7}

~\cite{qin2018compress}                & \textemdash                 & \begin{tabular}[c]{@{}l@{}}To provides a better selection mechanism \\ for deployment of DNN model\end{tabular}           & Guided deployment for wiser pruning                                                                &                       \cmark & \cmark & \cmark & \multicolumn{1}{c|}{}                                                                                                                                                             \\ \cline{1-7}

\cite{chauhan2018performance}         & \textemdash                  & \begin{tabular}[c]{@{}l@{}}Pruning method for deploying DNN model\\ on embedded devices\end{tabular}                      & \begin{tabular}[c]{@{}l@{}}Singular value decomposition based \\ factorization method\end{tabular}                       & \xmark & \xmark & \cmark & \multicolumn{1}{c|}{}                                                                                                                                                            \\ \cline{1-7}

~\cite{jacob2018quantization} & \textemdash                      & \begin{tabular}[c]{@{}l@{}}Inference is performed using integer\\ arithmetic\end{tabular}                   & \begin{tabular}[c]{@{}l@{}}Quantization technique is an affine \\ mapping\end{tabular}                                                       & \cmark & \cmark & \cmark & \multicolumn{1}{c|}{}                                                                                                                                                          \\ \hline
%%%%%%%%%%%%%%%%%%%%%%%%%%%%%%%%%%%%%%%%%%%%%%%%%%%%%%%%%%%%%%%%%%%%%%%%%%%%%%%%%%%%%%%%
~\cite{han2016eie}                     & EIE                                        & \begin{tabular}[c]{@{}l@{}}To perform DNN based inference on the \\ compressed network\end{tabular}                       & \begin{tabular}[c]{@{}l@{}}Pruning of redundant connections  and \\ weight sharing among different elements\end{tabular} & \cmark & \cmark & \cmark & \multicolumn{1}{c|}{}                                                                                                                                                             \\ \cline{1-7}

~\cite{eshratifar2020run}     & \textemdash                           & \begin{tabular}[c]{@{}l@{}}To jointly determine the model that should \\be called  to perform inference\end{tabular}                            & Developed lightweight neural multiplexer                                                                                                     & \cmark & \cmark & \cmark & \multicolumn{1}{c|}{\multirow{3}{*}{Multiplexing}}                                                                                                                             \\ \cline{1-7}

~\cite{zhang2019mbfn}         & \textemdash                & \begin{tabular}[c]{@{}l@{}}Multiplexing different recognition and \\prediction task using a single backbone network\end{tabular} & \begin{tabular}[c]{@{}l@{}}Sequential steps of training a CNN based \\ backbone network followed by branches\end{tabular} & \cmark & \xmark & \xmark & \multicolumn{1}{c|}{}                                                                                                                                                         \\ \cline{1-8}
%%%%%%%%%%%%%%%%%%%%%%%%%%%%%%%%%%%%%%%%%%%%%%%%%%%%%%%%%%%%%%%%%%%%%%%%%%%%%%%%%%%%%%%%%%

~\cite{han2015deep}           & \textemdash                & \begin{tabular}[c]{@{}l@{}}To identify the shared weights of each layer in \\ trained network\end{tabular}                                  & \begin{tabular}[c]{@{}l@{}}k-means clustering technique is used  where,\\ the same cluster must share the same weights\end{tabular}          & \cmark & \cmark & \cmark & \multicolumn{1}{c|}{\multirow{9}{*}{\begin{tabular}[c]{@{}c@{}}Weight\\ sharing\end{tabular}}}                                                                                  \\ \cline{1-7}

~\cite{han2016eie}                     & EIE                                        & \begin{tabular}[c]{@{}l@{}}To perform DNN based inference on the \\ compressed network\end{tabular}                       & \begin{tabular}[c]{@{}l@{}}Pruning of redundant connections  and \\ weight sharing among different elements\end{tabular} & \cmark & \cmark & \cmark & \multicolumn{1}{c|}{}                                                                                                                                                             \\ \cline{1-7}

~\cite{he2018multi}                    & MTZ                          & \begin{tabular}[c]{@{}l@{}}Minimal change in error upon layer sharing\end{tabular}                        & \begin{tabular}[c]{@{}l@{}}Layer pruning and weight update\end{tabular}                        & \cmark & \cmark & \cmark & \multicolumn{1}{c|}{}
\\ \cline{1-7}

~\cite{wu2018sharing}         & \textemdash                 & Data sharing during DNN model training                                                                                                          & \begin{tabular}[c]{@{}l@{}}Uses max-margin approach that extracts most \\ identifiable training data.\end{tabular}                           & \cmark & \cmark & \cmark & \multicolumn{1}{c|}{}                                                                                                                                                        \\ \cline{1-7}

~\cite{georgiev2017low}       & \textemdash                 & DNN based modeling framework for RCDs                                                                                                           & \begin{tabular}[c]{@{}l@{}}Framework follows a multitasking learning\\ principle for training shared DNN model\end{tabular}                 &      \cmark                 &    \cmark                   &    \cmark                   & \multicolumn{1}{c|}{}                                                                                                                                                            \\ \hline
%%%%%%%%%%%%%%%%%%%%%%%%%%%%%%%%%%%%%%%%%%%%%%%%%%%%%%%%%%%%%%%%%%%%%%%%%%%%%%%%%%%%%%%%%%%%%%%%%%%%%%%%%%%%%%%%%%%%%
\end{tabular}
}
\label{table3}
\end{table*}

\subsection{Quantization}
This section covers the sparse representation technique that involves weight quantization in the DNN model~\cite{han2015deep,qin2018compress,jacob2018quantization,chauhan2018performance}. The weight quantization reduces the storage requirement of the DNN model, along with the computation requirement. Authors in~\cite{han2015deep} proposed a weight quantization technique to compress the deep neural network by curtailing the number of bits required for representing the weight matrices. Here, authors try to reduce the number of weights that should store in the memory. In doing so, the same weights are eliminated, and multiple connections are drawn from a single remaining weight. Let us consider a scenario, where, a deep learning model has $4$ neurons at the input and output layers. Next, the weights are quantized to $4$ bins, where, all weights in the same bin share the same value. Thus, during the deployment of the DNN model on the embedded device,  we have to store smaller indices for each weight, as illustrated in part (a) of Fig.~$4$. Similarly, the gradients are also updated during the backpropagation of the DNN model and weights are reduced, as illustrated in part (b) of Fig.~$4$.

\begin{figure}
 \centering
 \includegraphics[scale=1.00]{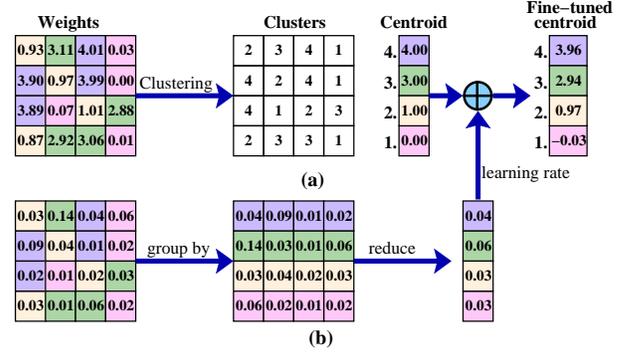}
 \caption{Illustration of weight quantization using clustering~\cite{han2015deep}. Part (a) Forward propagation and part (b) backward propagation.}
\end{figure}

Jacob \textit{et. al.}~\cite{jacob2018quantization} proposed a quantization technique, where, the inference is performed using integer arithmetic. The integer arithmetic provides higher efficiency than floating-point operation and requires a lower number of bits for representation. Authors further design a training step that preserves the accuracy compromise during replacement of floating-point operation with integer operations. The proposed approach thus solve the tradeoff between on-device latency and accuracy compromise due to integer operations. The authors have used integer arithmetic during inference and floating-point operation during training. The quantization technique is an affine mapping of integers $Q$ to the real number $R$, \textit{i.e.,} of the form
\begin{equation}
 R=W(Q-T),
 \label{affine}
\end{equation}
where, Eq.~\ref{affine} is the quantization scheme having quantization parameters W and T. For example, for $8$-bit quantization $Q$ is set to $8$. The quantization parameter $W$ is an arbitrary positive real number and $T$ is of the same type as variable $Q$.

\noindent $\bullet$ \textbf{Remarks:} The quantization technique discussed in existing literature~\cite{han2015deep,qin2018compress,jacob2018quantization,chauhan2018performance} for compressing DNN model. The techniques cover the model reduction by providing optimal arrangements of weight matrices. However, the negative consequences of weight quantization and its complexity of estimation are missing for the existing work.

\subsection{Multiplexing}
The multiplexing plays a vital role in reducing the storage requirement of the DNN model. The multiplexing is highly effective when weight matrices have similar values of weights~\cite{kadetotad2016efficient}. These weights are replaced with a single weight with multiplexed connections. The overview of the existing literature on multiplexing for DNN compression~\cite{han2016eie, eshratifar2020run, zhang2019mbfn} are as follows. Authors in~\cite{eshratifar2020run} proposed a lightweight neural multiplexer. The input to the multiplexer is the raw data and resource budget, which jointly determine the model that should be called to perform inference. The lightweight data with a low budget of resources are inferences on the mobile device, whereas, hard data with high resource budget resources inferences at the Cloud. In the multiplexing approach, multiple models are multiplexed from small to large. The proposed approach is output to a binary vector that indicates the inference is either on the mobile device or Cloud. The authors introduced a contrastive loss function ($Loss_{c}$) to train all the model in the multiplexing by adding $Loss_{c}$ with the main loss function (\textit{e.g. cross-entropy loss}) of the model. For a pair of models ($M_1$ and $M_2$), three possible cases can happen, \textit{i.e.,} both $M_1$ and $M_2$ can predict correct output, $M_1$ or $M_2$ can predict correct output, and non of them can predict correct output. In the third case, $Loss_{c}$ is not applied, and cross-entropy function only work. The contrastive loss function ($Loss_{c}$) is defined as

\begin{align}\nonumber
 Loss_{c}(\hat{y},y) =& \sum_{i=1}^{N}\sum_{j=1,i\ne j}^{N} \log \Big(dist(a_i,a_j)\Big)\\ \nonumber
 &\times \Big((\hat{y}_i==y\&\hat{y}_j==y)\\ \nonumber
 &-(\hat{y}_i!==y\&\hat{y}_j==y)\\
 &-(\hat{y}_i==y\&\hat{y}_j==y)\Big),
\end{align}
where, $y$ and $\hat{y}$ are correct and predicted labels, respectively. $dist{\cdot}$ denote the cosine distance function given by
\begin{equation}
 dist(a_1,a_2)=\frac{a_1^Ta_2}{a_1^Ta_1\times a_2^T a_2}.
\end{equation}

In~\cite{zhang2019mbfn} authors proposed a multi-branch CNN architecture, which multiplexes different recognition and prediction task simultaneously using a single backbone network. The proposed mechanism involves sequential steps, \textit{i.e.,} first, the authors have trained a CNN based backbone network then train different branches of the network by freezing the training of the backbone network. The authors proved the multiplex approach that preserves both time and energy of the system while achieving significant accuracy on an embedded device.

\noindent $\bullet$ \textbf{Remarks:} The existing literature on multiplexing~\cite{han2016eie, eshratifar2020run, zhang2019mbfn} helps in DNN compression for reducing storage and computation requirements. Though the existing literature has designed the framework for multiplexing different DNN models. However, none of the work tried to multiplex the weights, which can be an effective solution for reducing the storage of the DNN model.

\subsection{Weight sharing}
In this section, we cover the overview of the weight sharing in the DNN model. Weight sharing proves to be the important criteria for reducing both storage and computation requirements of the DNN model~\cite{larsen2020weight}. In weight sharing, inspite of storing multiple weights for the DNN model, it would be convenient to share the pre-existing weights. Here, the weights stored for $(i-1)^{th}$ layers can be used by the $i^{th}$ layer. The detail descriptions of weight sharing are as follows. Authors in~\cite{han2015deep} use k-means clustering technique for identifying the shared weights of each layer in the trained network. All the connections that are in the same cluster must share the same weights. Here, the authors also assumed that the weights are not shared across the layers in the DNN model. They partition $m$ weights in the network, $\mathbf{W}=\{w_1, w_2, \cdots w_m\}$ into $c$ clusters $K=\{k_1, k_2,\cdots k_c$\}, where, $m>>c$.

\begin{equation}
 \arg \min_{K} \sum_{i=1}^{C} \sum_{w\in k_i} |w-k_i|^2.
\end{equation}

Wu \textit{et. al}~\cite{wu2018sharing} emphasized the technique of sharing training process that involves sharing of training data. This sharing of data (or weights) provides a better understanding of the prediction process involve in the DNN model. However, this data sharing during the model training also leads to privacy compromise. Therefore, the authors proposed a method that discloses a few samples of the training data, which are sufficient for a data analyst to get insight into the DNN model. To reduce the discloser of the training data, authors have used the max-margin approach that extracts most identifiable training data. These identifiable data significantly contributes to obtaining a decision boundary. 

In~\cite{georgiev2017low} authors proposed deep learning-based modeling and optimization framework for resource constraint devices. The proposed framework achieves considerable accuracy while consuming limited resources. The framework follows a multitasking learning principle for training shared DNN model. Here, the hidden layers are shared among each other, where, similar weights are associated with multiple links after elimination. The presented framework balances the improvement in the performance by optimizing different losses in the multitasking framework. In the case of multitask learning, $T_s$ supervised tasks are considered having training dataset, $\mathcal{D}_i=(\mathbf{x}_j^i,y_j^i)_{j=1}^N$, where, $i \in \{1,2, \cdots, T_s\}$. Let $\mathcal{L}(\cdot)$ is the loss incurred during the training, the objective of the multi-tasking learning is to minimize following term

\begin{equation}
 \min \sum_{i=1}^{T_s} \frac{1}{N} \sum_{j=1}^{N} \mathcal{L}(y_j^{i},\phi(\mathbf{x}_j^{i})) + \lambda||\theta||^2.
\end{equation}

\noindent $\bullet$ \textbf{Remarks:} The existing literature on weight sharing~\cite{han2015deep,han2016eie,he2018multi,wu2018sharing,georgiev2017low} helps in reducing storage and computation requirement of the DNN model. However, the complexity of weight sharing also comes into the picture, while, representing DNN model using minimal weights. Thus, it could be beneficial to reduce the complexity of weight sharing in the DNN model.

\section{Bits precision}\label{bp}
This section presents an overview of the DNN compression technique that incorporates bits precision for reducing storage and computation requirements. In bits precision, the number of bits required for representing weight matrices is suppressed for reducing the storage and computation~\cite{jain2018compensated}. For example, we require $32$ bits for storing the weights in the weight matrix in floating-point operations of DNN model. It can be reduced to $8$ bits by performing operations using integers. This transformation from float-to-integer simultaneously reduces storage and computation requirements. However, it comes up with the challenge of conversion complexity from float to integer along with a higher accuracy compromise. The existing literature on bits precision~\cite{jacob2018quantization,lee2019neuro,hubara2017quantized,hubara2016binarized,yang2019cdeeparch,umuroglu2017finn,louizos2017bayesian} tried to tackle out these challenges. Table~\ref{table4} summarizes different DNN compression techniques using bits precision. The overview of the bits precision techniques in existing literature are as follows.

\begin{table*}[h]
\caption{Summary of the DNN compression incorporating bits precision for reducing computation and storage.}
\resizebox{1.0\textwidth}{!}{
\begin{tabular}{|c|c|l|l|l|l|l|l|}
\hline
\multirow{2}{*}{\textbf{Paper}}                              & \multirow{2}{*}{\textbf{\begin{tabular}[c]{@{}c@{}}Abbreviated \\ name\end{tabular}}}  & \multicolumn{1}{c|}{\multirow{2}{*}{\textbf{Address the challenge}}}                                                      & \multicolumn{1}{c|}{\multirow{2}{*}{\textbf{Proposed solution}}}                                                         & \multicolumn{3}{c|}{\textbf{Suitable for}}                            & \multirow{2}{*}{\textbf{Category}}                                                                                                                                                     \\ \cline{5-7} 
                                                             &                                            & \multicolumn{1}{c|}{}                                                                                                     & \multicolumn{1}{c|}{}                                                                                                    & \textbf{CNN}          & \textbf{FC}           & \textbf{RNN}          &                                                              \\ \hline
~\cite{lee2019neuro}          & Neuro.ZERO                  & \begin{tabular}[c]{@{}l@{}}Adopt integer arithmetic by replacing \\ floating-point operations\end{tabular}                                      & \begin{tabular}[c]{@{}l@{}}Co-processor architecture for toggling during\\  availability and scarcity of resources\end{tabular}              & \cmark & \cmark & \cmark & \multicolumn{1}{c|}{\multirow{2}{*}{\begin{tabular}[c]{@{}c@{}}Estimation\\ using integer\end{tabular}}}  \\ \cline{1-7}

~\cite{jacob2018quantization}                   & \textemdash                                                            & To reduce the temporary storage for computation                                                                                                 & Quantization mechanism that uses integer            & \cmark & \cmark & \cmark & \multicolumn{1}{c|}{}                                                                                                                                                               \\ \hline
%%%%%%%%%%%%%%%%%%%%%%%%%%%%%%%%%%%%%%%%%%%%%%%%%%%%%%%%%%%%%%%%%%%%%%%%%%%%%%%%%%%%%%%%%%%%%%%%%%%%%%%%%%%%%%
\cite{hubara2017quantized}                           & QNN        & \begin{tabular}[c]{@{}l@{}}To design a DNN network that requires very low \\ precision weight and activation during inference\end{tabular}     & \begin{tabular}[c]{@{}l@{}}Linear and logarithmic scheme for reducing\\ the bits requirements\end{tabular}                                  & \cmark & \cmark & \cmark & \multicolumn{1}{c|}{\begin{tabular}[c]{@{}c@{}}Low bits\\ representation\end{tabular}}                                                                                               \\ \hline
%%%%%%%%%%%%%%%%%%%%%%%%%%%%%%%%%%%%%%%%%%%%%%%%%%%%%%%%%%%%%%%%%%%%%%%%%%%%%%%%%%%%%%%%%%%%%%%%%%%%%%%%%%%%%%%%%%
~\cite{hubara2016binarized}                     & BNN           & \begin{tabular}[c]{@{}l@{}}To replace floating-point operations in the DNN\\ with binary operations\end{tabular}                                & Deterministic and stochastic techniques                                                                                          & \cmark & \cmark & \cmark & \multicolumn{1}{c|}{\multirow{5}{*}{Binarization}}                                                                                                                                  \\ \cline{1-7}

~\cite{umuroglu2017finn}                        & FINN          & \begin{tabular}[c]{@{}l@{}}Heterogeneous architecture for achieving high \\ order flexibility\end{tabular}                                      & \begin{tabular}[c]{@{}l@{}}Parametric architecture for dataflow \\and optimized method for classification\end{tabular}                      & \cmark & \cmark & \cmark & \multicolumn{1}{c|}{}                                                                                                                                                                \\ \cline{1-7}

~\cite{yang2019cdeeparch} & cDeepArch    & \begin{tabular}[c]{@{}l@{}}Dividing the task into multiple sub-task and \\using binarization of weights\end{tabular}                            & \begin{tabular}[c]{@{}l@{}}Quantitative measure to estimate the \\reduction in resources \end{tabular}   & \cmark & \xmark & \xmark & \multicolumn{1}{c|}{}                                                                                       
                                                                        \\ \hline
\end{tabular}
}
\label{table4}
\end{table*}

\subsection{Estimation using integer}
This section covers, the most straightforward strategy of reducing computation through bits precision is replacing floating-point operations with the integers. Here, the only complexity is to convert floating values to the integer. The existing work on bits precision using integer~\cite{jacob2018quantization,lee2019neuro} are elaborated as follows. Authors in~\cite{lee2019neuro} proposed a mechanism of opportunistically accelerating the inference performance of the DNN model, named Neuro.ZERO. They adopt four accelerating mechanisms, \textit{i.e.,} extended inference, expedited inference, ensemble inference, and latent training. Further, the authors proposed two algorithms for applying these accelerating mechanisms. They have adopted integer arithmetic by replacing floating-point operations. The authors emphasized on facilitating runtime adaptations, where, accuracy during runtime suppose to increase when the resources are available. The authors adopt co-processor architecture for toggling during availability and scarcity of resources. The extended inference involves an extension of DNN structure that improves the accuracy. It uses additional resources of the secondary processor for running extended part Next, the expedited inference speedup by offloading some task portion of the original DNN model. Later, in ensemble inference, multiple DNN models run simultaneously on primary and secondary processors, and their outputs are combined. Finally, in latent training primary model run as usual, but for unseen data, training is performed on secondary processors.

\begin{figure}[h]
 \centering
 \includegraphics[scale=1.07]{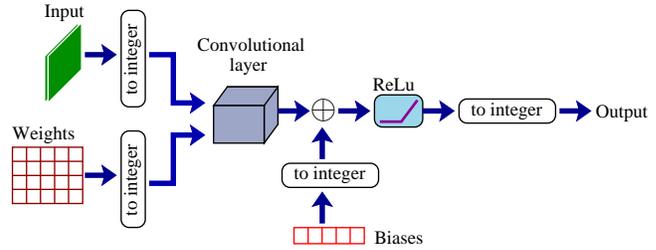}
 \caption{Process of converting floating point operations into integer operations in DNN model.}
 \label{integer}
\end{figure}

Jacob \textit{et. al.}~\cite{jacob2018quantization} proposed a quantization mechanism, which uses integer arithmetic despite floating points operations. It relies on the facts that multiplications are inefficient if the operands are wide (floating points $32$ bits) and eliminating multiplication leads to performance degradation. Therefore, it could be beneficial to adopt integer multiplication inspite of floating-point operations to reduce the temporary storage for computation. Fig.~\ref{integer} illustrates the entire steps involve in performing integer operations by replacing all floating point operations. 

\noindent $\bullet$ \textbf{Remarks:} The estimation using the integer values plays a vital role in reducing the bits requirements for storing weights. The prior studies~\cite{jacob2018quantization,lee2019neuro} have successfully performed the DNN operations using integer values. However,  they lack in estimating the amount of memory we can preserve through integer estimation. Additionally, beyond integer, other low bits estimations are also possible, which could be explored.

\subsection{Low bits representation}
This section extends the concept of the integer representation to reduce the storage and computation of DNN model. It generalizes the concept of bits precision for using any number of bits for representing weights instead of $8$-bits integers only. The concept of Quantized Neural Networks (QNNs) is discussed in~\cite{hubara2017quantized}. Here, the authors design a DNN network that requires very low precision (\textit{e.g.,} $1$ bits, $4$ bits, and so on) weight and activations during inference. While training the quantized weights and activations helps in estimating the gradient. This QNN scheme highly reduces memory requirements, access latency, and replace floating-point operations with bit-wise operations. Therefore, colossal power efficiency is observed. Authors have used two types of quantization schemes for reducing the bits requirements for representing weights and activations, namely, linear and logarithmic. The linear quantization $L_{Q}(\cdot)$ is defined as 

\begin{align}
L_{Q}(r,l_b)=Clip\Big(round\Big(\frac{r}{2^{l_b}-1}\Big)2^{l_b-1},V_{\min}, V_{\max}\Big),
\end{align} 
where, $r$ is the real valued number, $l_b$ is the length of bits, and $round(\cdot)$ is the round-off function. $V_{\min}$ and $V_{max}$ are the minimum and maximum scale range, respectively. Further, the logarithmic quantization $Log_{Q}(\cdot)$ is given as follows

\begin{align}
Log_{Q}(r,l_b)=Clip\Big(\log ap2 (r), -(2^{l_b-1}-1), 2^{l_b-1}\Big),
\end{align} 
where, $\log ap2(r)$ denotes the log of approximate power of $2$, in other words it uses the logarithm for most significant bits.

\noindent $\bullet$ \textbf{Remarks:} The low bits representation in~\cite{hubara2017quantized} is an extension of integer estimation and provides the facility of using any number of bits for performing the DNN operations. However, selecting an optimal number of bits for any estimation is a tedious task.

\subsection{Binarization}
In this section, we illustrate an overview of bits precision techniques~\cite{hubara2016binarized, umuroglu2017finn, yang2019cdeeparch} that incorporates binary number for performing the different operation of the DNN model. Binarization refers to the process of converting floating-point operations into binary operations for reducing the storage and processing requirement of the DNN model~\cite{tang2019towards}. The overview of the different binarization techniques in DNN models is as follows. Authors in~\cite{hubara2016binarized} introduced a training method that incorporates binary activations and weights. The resultant network is called Binary Neural Network (BNN). During the training of the model; the gradient is estimated using binary activations and weights. Binarization helps in reducing storage requirements and provides energy efficiency. Here, the floating-point operations in the DNN model are replaced with binary operations. The authors in the paper have highlighted two binarization techniques, \textit{i.e.,} deterministic and stochastic. In deterministic binarization, the real-valued number $r$ is transformed into binary value $b$ using $Sign(\cdot)$ function, which is defined as

\begin{align}
 b=Sign(r)=\left\{\begin{matrix}
+1 & \text{if } r\ge0,  \\ 
-1 & \text{otherwise}.
\end{matrix}\right.
\end{align} 

In stochastic binarization, a probability $p$ is estimated for converting real value $r$ to binary value $b$ as

\begin{align}
 b=\left\{\begin{matrix}
+1 & \text{if } p = \max \Big(0,\min\Big(1,\frac{r+1}{2}\Big)\Big),  \\ 
-1 & \text{if } 1-p.
\end{matrix}\right.
\end{align} 

Umuroglu \textit{et. al.}~\cite{umuroglu2017finn} presented a framework for building a fast and flexible DNN compression mechanism, named FINN. It provides a heterogeneous architecture that achieves high order flexibility. The framework is a parametric architecture for dataflow and optimized method for classification on limited resource device. The authors have implemented fully connected, convolutional, and pooling layers with computation as per the requirements of the user. It provides compressed DNN with sub-microsecond latency, thus enhance its suitability for deployment on embedded devices.

In~\cite{yang2019cdeeparch} authors presented a compact DNN architecture, where a task is divided into multiple subtasks for reducing the resource requirement of the DNN model. The authors named the approach as cDeepArch. The decomposition of the task in cDeepArch efficiently utilizes the limited storage and processing capacity during the execution of the task. The authors emphasized that the unorganized compression of the DNN models results in unreliable and low performance for recognizing multiple tasks. Further, the authors established a quantitative measure to estimate the reduction in resources and available resources. cDeepArch incorporates offline training followed by the execution of the task on user smartphone. It also preserves the privacy of sensitive information. Here, to provide privacy of the sensitive information, it is converted into image format for training a DNN model. For reducing the DNN model, cDeepArch adopted layer separation, delayed pooling, integration of activation, and binarization for reducing the computation.

\noindent $\bullet$ \textbf{Remarks:} The binarization of DNN model presented in the prior studies~\cite{hubara2016binarized, umuroglu2017finn, yang2019cdeeparch} reduce the storage and computations. The proposed binarization approaches achieve a high-order compression but with performance degradation. Thus, it could be beneficial to mitigate this degradation for the successful adaptation of binarization techniques in DNN compression.

\section{Knowledge distillation}\label{kd}
This section gives an overview of the DNN compression technique that adopts knowledge distillation~\cite{phuong2019towards} to improve the performance of the compressed DNN model. The existing literature on knowledge distillation~\cite{wang2019private,hinton2015distilling,yun2020regularizing,li2020few,cheng2020explaining, mirzadeh2019improved,duong2019shrinkteanet,yang2020mobileda} is further classified into three parts, \textit{i.e.,} logits transfer, teacher assistant, and domain adaptation, as illustrated in Table~\ref{table5}. In logits transfer, the unnormalized output of the original DNN model (teacher) act as a soft target for training the compressed DNN model (student)~\cite{abbasi2020modeling}. However, sometimes the gap between student and teacher is large, which hinders the appropriate knowledge transmission.  It could be mitigated by inserting a teacher assailant in the between teacher and student. Further, the domain adaptation is discussed, where, the generalization ability of the teacher model could not improve the performance of the student model due to domain disparity between training and testing data of the student model.

\begin{table*}[h]
\caption{Illustration of knowledge distillation techniques for improving the performance of compressed DNN model.}
\resizebox{1.0\textwidth}{!}{
\begin{tabular}{|c|c|l|l|l|l|l|l|}
\hline
\multirow{2}{*}{\textbf{Paper}}                              & \multirow{2}{*}{\textbf{\begin{tabular}[c]{@{}c@{}}Abbreviated \\ name\end{tabular}}}  & \multicolumn{1}{c|}{\multirow{2}{*}{\textbf{Address the challenge}}}                                                      & \multicolumn{1}{c|}{\multirow{2}{*}{\textbf{Proposed solution}}}                                                         & \multicolumn{3}{c|}{\textbf{Suitable for}}                            & \multirow{2}{*}{\textbf{Category}}                                                                                                                                                     \\ \cline{5-7} 
                                                             &                                            & \multicolumn{1}{c|}{}                                                                                                     & \multicolumn{1}{c|}{}                                                                                                    & \textbf{CNN}          & \textbf{FC}           & \textbf{RNN}          &                                                              \\ \hline
                                                             
~\cite{hinton2015distilling}  & \textemdash    & \begin{tabular}[c]{@{}l@{}}Improving performance of compression \\ DNN model\end{tabular}                                & \begin{tabular}[c]{@{}l@{}}Knowledge distillation technique \\ for improving performance\end{tabular} & \cmark & \cmark & \cmark & \multicolumn{1}{c|}{\multirow{11}{*}{\begin{tabular}[c]{@{}c@{}}Logits \\ transfer\end{tabular}}}                      \\ \cline{1-7}

~\cite{yun2020regularizing}   & \textemdash   & \begin{tabular}[c]{@{}l@{}}Distilling the predictive distribution \\ among same class labels for training\end{tabular}   & \begin{tabular}[c]{@{}l@{}}Performance using self-knowledge \\ distillation\end{tabular}  & \cmark & \cmark & \cmark &                                                                                                                                                                                 \\ \cline{1-7}

~\cite{wang2019private}       & RONA      & \begin{tabular}[c]{@{}l@{}}DNN compression to provide \\ significant privacy to the user\end{tabular}                    & \begin{tabular}[c]{@{}l@{}}Designed a private model \\ compression framework\end{tabular}    & \cmark & \cmark & \cmark &                                                                                                                                                                                 \\ \cline{1-7}

~\cite{li2020few}             & FSKD     & \begin{tabular}[c]{@{}l@{}}To highlight the complexity of \\ fine-tuning of compress DNN model\end{tabular}              & \begin{tabular}[c]{@{}l@{}}Established a relationship between \\ compression rate and training epochs\end{tabular}      & \cmark & \cmark & \cmark &                                                                                                                                                                                   \\ \cline{1-7}

~\cite{cheng2020explaining}   & \textemdash       & Interpretation of knowledge distillation                                                                                 & \begin{tabular}[c]{@{}l@{}}Involve both relevant and irrelevant \\ features related to the visual task\end{tabular}     & \cmark & \cmark & \cmark &                                                                                                                                                                                  \\ \hline
%%%%%%%%%%%%%%%%%%%%%%%%%%%%%%%%%%%%%%%%%%%%%%%%%%
~\cite{mirzadeh2019improved}  & \textemdash   & \begin{tabular}[c]{@{}l@{}}To reduce the gap between student \\ and teacher model\end{tabular}                           & \begin{tabular}[c]{@{}l@{}}Introduced teacher assistant between \\ teacher and student during distillation\end{tabular} & \cmark & \cmark & \cmark & \multicolumn{1}{c|}{\begin{tabular}[c]{@{}c@{}}Teaching\\ assistant\end{tabular} }                                                                                                                  \\ \hline

%%%%%%%%%%%%%%%%%%%%%%%%%%%%%%%%%%%%%%%%%%%%%%%%%%%%%%%%%
~\cite{duong2019shrinkteanet} & ShrinkTeaNet    & \begin{tabular}[c]{@{}l@{}}To train a few parameter student models\\ using cumbersome teacher model\end{tabular}         & \begin{tabular}[c]{@{}l@{}}Developing a technique that is more \\ robust towards open-set problem\end{tabular}          & \cmark & \xmark & \xmark & \multicolumn{1}{c|}{\multirow{3}{*}{\begin{tabular}[c]{@{}c@{}}Domain\\ distillation\end{tabular}}}  \\             \cline{1-7}

~\cite{yang2020mobileda}      & MobileDA         & \begin{tabular}[c]{@{}l@{}}To perform training on simplified DNN \\ model that handles domain shift problem\end{tabular} & \begin{tabular}[c]{@{}l@{}}Learning transferable features for\\  domain adaptation\end{tabular}     & \cmark & \xmark & \xmark &                                                                                                                                                                                  \\ \hline

\end{tabular}
}
\label{table5}
\end{table*}

\subsection{Logits transfer}
Logits transfer is the simplified approach for knowledge distillation from teacher to student model. Firstly, the teacher model is trained on a given dataset $\mathcal{D}$. Next, the logit vectors (output of DNN model before the softmax layer) of the teacher model acts as a soft target for a training student model. The knowledge distillation using logits transfer improves the generalization ability of the student model and helps in improving its performance. Different logits transfer mechanism in existing literature are described as follows. 

Authors in~\cite{hinton2015distilling} proposed a knowledge distillation technique for improving the performance of a compressed DNN model using the generalization ability of the pre-trained cumbersome model. The cumbersome model contains the ensembled output of multiple models, thus achieve higher generalization ability. The deployment of the cumbersome model on an embedded device is impractical due to substantial resource requirements. To overcome such limitations, the small model is deployed on the embedded device, with accuracy improvement using the cumbersome model. The principle objective is to distil the knowledge from large model to small model by training small model in such a way that it achieves similar performance as the cumbersome model. While performing the transmission of the generalization ability of the cumbersome model to the small compressed model, the class label probabilities will act as a soft target for training the small model.

In the DNN model, the output features vector obtained at one layer prior to the softmax layer is termed as logit. Let $\mathbf{a}_i$ denotes a logit vector of data instance $i$ $(\forall i \in N)$, where, $N$ represents total number of data instances in dataset $\mathcal{D}$. To obtain $\mathbf{a}_i$, let $x_{ij}\in \mathbf{X}$, $w_{ij}\in \mathbf{W}^T $, and $b_{j}\in \mathbf{b}$ represent an element of feature matrix, weight matrix, and bias vector for $j^{th}$ ($1\leq j \leq l$) class of $i^{th}$ ($1\leq i \leq N$) training instance, respectively, where, $l$ denotes the number of classes in $\mathcal{D}$. Hence, we can estimate an element $a_{ij}$ of logit vector $\mathbf{a}_i$ for class $j$ given as
 
\begin{equation}
a_{ij}=w_{ij} x_{ij} + b_{j}.
\end{equation}

Later, the logits vector $\mathbf{a}_i =\{a_{ij} | 1\leq j \leq l\}$ passes to a softmax function to compute predicted class label probability $p_{ij}$ as following
\begin{equation}\label{pp}
 p_{ij}=\frac{e^{a_{ij}}}{\sum_{j=1}^{l}e^{a_{ij}}}.
\end{equation}

The predicted probability vector for $i^{th}$ instance is a set of probabilities \{$p_{i1},p_{i2},\cdots,p_{il}$\}, against each class label $l$ and a class label with highest probability value is said to be the predicted class label. Further, we introduce a variable $\mathcal{T}$ for generating a softer probability distribution; therefore, Eq.~\ref{pp} is rewritten as

\begin{equation}\label{pp}
 \rho_{ij}=\frac{e^{u_{ij}/\tau}}{\sum_{j=1}^{k}e^{u_{ij}/\tau}}.
\end{equation}

The matching of the logits mainly involves two methods~\cite{chen2018distilling}, \textit{i.e.,} cross-entropy using soft targets and cross-entropy using correct labels. In cross-entropy using soft targets, the cross-entropy loss of compressed model is calculated using the large value of temperature $\mathscr{T}$, which was used by the cumbersome model for generating soft targets. Similarly, in cross-entropy with correct labels, we compute the cross-entropy loss of compressed model using similar logit as the cumbersome model. To perform the matching of logits, we have to estimate gradient $\nabla$ with the element ($a_{ij}$) of logit vector. $a_{ij}$ is the element of logit vector of $i^{th}$ data instance and $j^{th}$ class. Let $b_{ij}$ represent the element of logit vector of the cumbersome model, then the gradient $\nabla$ is estimated as

\begin{align} \nonumber
\nabla=& = \sum_{j=1}^{l}\frac{1}{\mathcal{T}}(p_{ij}-q_{ij}) \\ \label{maineq}
&= \sum_{j=1}^{l}\frac{1}{\mathcal{T}}\Big(\frac{e^{a_{ij}/\mathcal{T}}}{\sum_{j=1}^{l}e^{a_{ij}/\mathcal{T}}}-\frac{e^{b_{ij}/\mathcal{T}}}{\sum_{j=1}^{l}e^{b_{ij}/\mathcal{T}}}\Big).
\end{align}

At the higher value of $\mathcal{T}$ the magnitude of matching logits in Eq.~\ref{maineq} can be rewritten in an approximate format as follows

\begin{equation}\small
 \Delta \approx \sum_{j=1}^{l} \frac{1}{\mathcal{T}}\Big(\frac{1+a_{ij}/\mathcal{T}}{N+\sum_{j=1}^{l}a_{ij}/\mathcal{T}}-\frac{1+b_{ij}/\mathcal{T}}{N+\sum_{j=1}^{l}b_{ij}/\mathcal{T}}\Big).
 \label{final}
\end{equation}

\begin{figure}[h]
 \centering
 \includegraphics[scale=1.05]{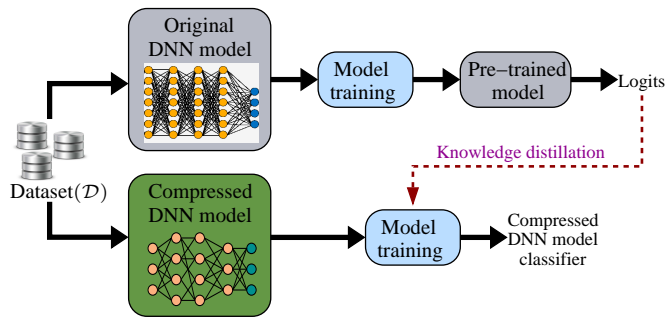}
 \caption{Illustration of knowledge distillation using logits of original DNN model.}
 \label{distillation_l}
\end{figure}

The main objective of knowledge distillation is to minimize the gradient ($\nabla$) in Eq.~\ref{final}. The value of $\nabla=0$ indicates the performance of the cumbersome and compressed DNN model is exactly same. However, it is impractical to reach this equilibrium state. As it is achieved when the compressed model is trained for an infinite number of epochs with logits of teacher model. To avoid such a problem, we stop training the compressed model, when $\nabla\rightarrow0$. At this stage, the compressed model achieves significant performance by consuming minimal resources. Fig.~\ref{distillation_l} illustrates the steps involve in the knowledge distillation using logits transfer.

Yun \textit{et. al.}~\cite{yun2020regularizing} distilled the predictive distribution among the same class labels for training. This distillation results in the regularization of knowledge obtained from wrong predictions, in a single DNN model. In other words, the model improves its performance using self-knowledge distillation by rapidly training and employing more precise predictions. Thus, the authors have tried to improve the performance of a build classifier in self-improving mode using dark knowledge of wrong predictions. The authors have combined two loss functions, \textit{i.e.,} cross-entropy and Kullback-Leibler (KL) divergence loss, using a regularization parameter $\lambda$. The combined loss function $\mathcal{L}_{comb}(\cdot)$ is defined as

\begin{equation} \nonumber
 \mathcal{L}_{comb}(\mathbf{x},\mathbf{x}{'},y,\phi,\mathcal{T})= \mathcal{L}_{CE}(\mathbf{x},y,\phi) + \lambda \mathcal{T}^2 \mathcal{L}_{KL}(\mathbf{x},\mathbf{x}{'}, \phi, \mathcal{T}),
\end{equation}
where, $\mathcal{L}_{CE}(\cdot)$ and $\mathcal{L}_{KL}(\cdot)$ are the cross-entropy and KL-divergence losses, respectively. $\mathbf{x}$ is the data instance instance having label $y$ and $\mathbf{x}{'}$ is another data instance having label $y$. $\phi$ is the network parameter and $\mathcal{T}$ is the temperature used in the knowledge distillation for softening the prediction probabilities of the knowledge inferring model.

In~\cite{wang2019private} authors have designed a private model compression framework, named pRivate mOdel compressioN frAmework (RONA). The authors have highlighted the need for DNN compression to provide significant privacy to the user by analysing the collected data on the limited resource devices. Using knowledge distillation, the accuracy of the compressed model is improved by jointly using a hint, distillation, and self learnings. In performing knowledge distillation, the cumbersome model is carefully perturbed for enforcing a high level of privacy. Therefore, the authors try to meet two crucial goals simultaneously, \textit{i.e.,} model compression and preserve privacy.

Authors in~\cite{li2020few} highlighted the complexity of fine-tuning of compress DNN model. The compression of DNN using pruning, quantization, weight decomposition requires fine-tuning, which may take a massive number of epochs for stabilization. Therefore, they established a relationship that higher the compression rate higher will be the epochs counts for stabilizing the performance of the model. Another problem associated with fine-tuning is the compressed DNN model that requires large training dataset for achieving considerable accuracy. Here, the authors try to realize both data and processing efficiency. The authors proposed a Few Sample Knowledge Distillation (FSKD) that performs efficient network compression. The efficiency is achieved in terms of both model training and inference.

In~\cite{cheng2020explaining} authors presented a method for successful interpretation of knowledge distillation that involve both relevant and irrelevant features related to the visual task. They presented three types of hypothesis for knowledge distillation, including, 1) it helps in learning more precise class label predictions from raw data, 2) it ensures simultaneously learning multiple class form the dataset, and 3) it provides the well-optimized direction for leaning. The authors have demonstrated the pros and cons of the hypothesis using empirical and experimental analysis on the different datasets.

\noindent $\bullet$ \textbf{Remarks:} The existing work on logits transfer~\cite{wang2019private,hinton2015distilling,yun2020regularizing,li2020few,cheng2020explaining} in knowledge distillation improves the performance of the compressed DNN model. However, the logits of original pre-trained DNN model sometimes lead to overfitting, which leads to poor generalization ability of the compressed model. As during fine-tuning of the compressed DNN model, the student model can mimic the exact prediction behaviour as teacher model but lose its generalization ability. It could be a research direction to determine the exact point for halting the training of the student model to retain its generalization ability and avoid overfitting.

\subsection{Teacher assistant}
\begin{figure}[h]
 \centering
 \includegraphics[scale=1.0]{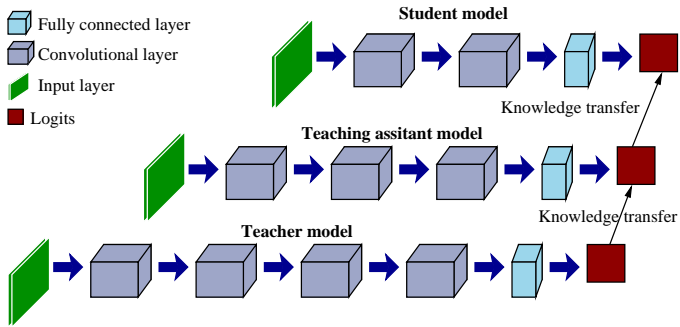}
 \caption{An example scenario of teacher assistant between teacher model and student model~\cite{mirzadeh2019improved}.}
 \label{t_assistant}
\end{figure}

Authors in~\cite{mirzadeh2019improved} studied the concept of knowledge distillation from a different perspective. The authors introduced teacher assistant in between teacher and student model during knowledge distillation. This insertion of teacher assistant has the main motive to reduce the gap between student and teacher model. They claim, when the size of the student model is fixed, then we can not employ a large teacher model. In other terms, the gap between teacher and student model beyond a limit leads to improper knowledge transmission from student to teacher. To mitigate such difficulty in improving the performance of the student model through knowledge distillation, the teacher assistant plays a decisive role. Fig.~\ref{t_assistant} illustrates an example scenario demonstrates the role of teacher assistant to fill the gap between large teacher model and small student model. The authors also suggest the futuristic approach of inserting multiple teaching assistants for providing high order compression with minimal accuracy compromise.

\noindent $\bullet$ \textbf{Remarks:} The work in~\cite{mirzadeh2019improved} have discussed the benefits of inserting a teaching assistant between teacher and student models, but unfortunately, it will increase the training time by nearly two-fold. First, the training of teacher assistant is performed using teacher model followed by training of student model using teacher assistant. Despite significant improvement in the student model performance, the complexity of training could be a loophole of the proposed approach.

\subsection{Domain adaptation}
\begin{figure}[h]
 \centering
 \includegraphics[scale=0.90]{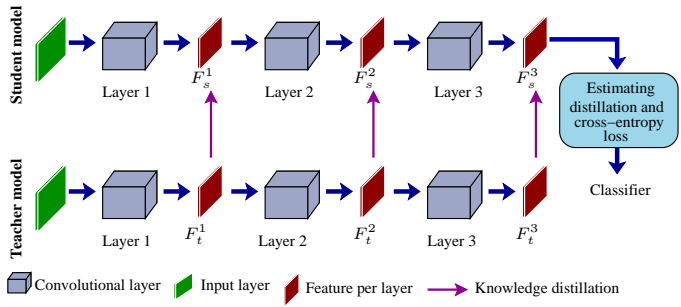}
 \caption{Layer-by-layer knowledge distillation from teacher to student for eliminating domain disparity.}
 \label{domain_t}
\end{figure}

The knowledge distillation techniques discussed in the previous sections do not consider the consequence of domain disparity while training student model using logits of the teacher. The domain disparity leads to poor generalization ability of the student model~\cite{ben2007analysis}. To handle domain disparity, different domain adaptation techniques have been proposed in the existing literature~\cite{wang2019private,duong2019shrinkteanet,yang2020mobileda}. Fig.~\ref{domain_t} illustrates the layer-by-layer feature extraction from teacher and student model followed by knowledge distillation from teacher to student.
The simultaneous training helps in preserving the student model generalization ability. 

The overview of different work on knowledge distillation incorporating domain adaptation are as follows.

Duong \textit{et. al.}~\cite{duong2019shrinkteanet} proposed a knowledge distillation technique that helps in shrinking the gap between cumbersome teacher and compact student model. Therefore, the authors called this technique as ShrinkTeaNet. The objective of ShrinkTeaNet is simple to train a few parameter student models using cumbersome teacher model. However, the authors emphasized on developing a technique that is more robust towards open-set problem along with maintaining significant accuracy. Further, authors have introduced angular distillation loss for determining the feature direction despite the hard targets that result in overfitting. The angular distillation loss helps in reducing the domain adaptation problem in knowledge distillation. The angular distillation loss ($ \mathcal{L}_{adl}(S,T)$) is defined as

\begin{align}\nonumber
 \mathcal{L}_{adl}(S,T) &= dist\Big(trans(F_t),trans(F_s)\Big)\\ \label{adl}
 &=\Big\| 1-\frac{trans(F_t)}{trans(F_t)}\times \frac{trans(F_s)}{trans(F_s)}\Big\|_2^2
\end{align}
where, $dist(\cdot)$ is the distance function that estimates the distance between the transformation function of the teacher ($trans(F_t)$) and student ($trans(F_s)$). Eq.~\ref{adl} is similar to the cosine similarity for estimating the distance between two vectors. In this approach, the authors suggest training the student model, along with the teacher model parallelly. This parallel training provides a soft target at each step despite final class label predictions that seem to be a hard target for the student model during training.

In~\cite{yang2020mobileda} authors emphasized on the problem of domain adaptation that is encountered during the training of the DNN model on embedded Internet of Things (IoT) devices. The authors discovered the DNN model achieves prominent result on the high-end machine but suffers from degradation in accuracy and efficiency on IoT devices. They proposed a framework for Mobile Domain Adaptation (MobileDA) that can learn the transferrable features. The learning is performed in such a way that the structure of the DNN model remains simplified as possible. The authors use cross-domain distillation for training student model on IoT devices using cumbersome teacher model on the high-end machine. The main objective of the MobileDA is to perform training on simplified DNN model that can handle the domain shift problem along with achieving significant accuracy. In other words, the training phase addresses the domain adaptation using cumbersome DNN model running on a high-end machine and testing phase is performed using a simplified DNN model deployed on the embedded device. Further, the authors proposed an algorithm that simultaneously optimizes different loss function for handling domain adaptation.

\noindent $\bullet$ \textbf{Remarks:} The domain adaptation technique proposed in the existing literature~\cite{wang2019private,duong2019shrinkteanet,yang2020mobileda} have successfully covered the domain disparity problem. However, the gap between the student model and teacher model sometimes hampers the performance of the domain adaptation technique. Thus, it could be beneficial to adopt the teacher assistant in domain adaptation problem to get overall benefits.

\section{Miscellaneous}\label{mln}
This section covers the DNN compression technique in the existing literature~\cite{yao2017deepsense, yao2018fastdeepiot, abc,radu2016towards,fang2018nestdnn,radu2018multimodal,mathur2017deepeye,xu2018deepcache,shen2019deepapp,xue2019deepfusion,svyatkovskiy2017training,wang2018deep,zhang2018shufflenet} that do not meet the different categorization criteria discussed in the previous sections. Apart from the outlier of categorization criteria, some of them simultaneously incorporate a group of DNN compression techniques. The principal motive behind the miscellaneous category, \textit{i.e.,} to reduce the storage and computation requirement of the DNN model is same as the above-discussed techniques. In other words, the miscellaneous category enlights the optimization of various DNN parameters before actually building the model. Thus, it helps in developing a storage and computationally efficient model for resource constraint devices. The overview of the DNN compression technique under miscellaneous category is illustrated in Table~\ref{table6}.

\begin{table*}[h]
\caption{Summary of miscellaneous approaches for compressing DNN model.}
\resizebox{1.0\textwidth}{!}{
\begin{tabular}{|c|c|l|l|l|l|l|}
\hline
\multirow{2}{*}{\textbf{Paper}} & \multirow{2}{*}{\textbf{\begin{tabular}[c]{@{}c@{}}Abbreviated \\ name\end{tabular}}} & \multicolumn{1}{c|}{\multirow{2}{*}{\textbf{Address the challenge}}}                                                                 & \multicolumn{1}{c|}{\multirow{2}{*}{\textbf{Proposed solution}}}                                                          & \multicolumn{3}{c|}{\textbf{Suitable for}} \\ \cline{5-7} 
                                &                                                                                       & \multicolumn{1}{c|}{}                                                                                                                & \multicolumn{1}{c|}{}                                                                                                     & \textbf{CNN}  & \textbf{FC} & \textbf{RNN} \\ \hline
~\cite{yao2017deepsense}                & DeepSense                                                                             & \begin{tabular}[c]{@{}l@{}}To automatically handle the noise in \\ the dataset collected form mobile device\end{tabular}             & \begin{tabular}[c]{@{}l@{}}Unified framework for on device \\ classification\end{tabular}                                 & \cmark         & \cmark       & \cmark        \\ \hline

~\cite{yao2018fastdeepiot}              & FastDeepIoT        & \begin{tabular}[c]{@{}l@{}}To exploits the non-linear relation between \\ structure of DNN model and its execution time\end{tabular} & \begin{tabular}[c]{@{}l@{}}Determined the tradeoff between execution \\ time and accuracy on embedded device\end{tabular} & \cmark         & \cmark       & \cmark        \\ \hline
~\cite{abc}                             & AdaDeep                                                                               & \begin{tabular}[c]{@{}l@{}}To automatically specifies a combination of \\ compression techniques for DNN model\end{tabular}          & \begin{tabular}[c]{@{}l@{}}Determine an optimal balance between \\ resources and user demanded performance\end{tabular}   & \cmark         & \cmark       & \cmark        \\ \hline

~\cite{radu2016towards}                 & \textemdash                                                                            & \begin{tabular}[c]{@{}l@{}}To determine the capability of DNN model \\ to perform recognition\end{tabular}                           & \begin{tabular}[c]{@{}l@{}}To utilize cross-sensor correlation for providing \\ a high-order of recognition performance\end{tabular} & \cmark         & \xmark       & \xmark        \\ \hline

~\cite{fang2018nestdnn}                 & NestDNN              & \begin{tabular}[c]{@{}l@{}}Emphasized on the availability of dynamic \\ resources on mobile devices\end{tabular}                     & \begin{tabular}[c]{@{}l@{}}To select optimal resources while performing \\ inference on the mobile device\end{tabular}               & \cmark         & \cmark       & \cmark        \\ \hline

~\cite{radu2018multimodal}              & \textemdash   & \begin{tabular}[c]{@{}l@{}}Benefits of using multimodal-sensors for \\ recognition using DNN on embedded devices\end{tabular}        & Using the modality-specific partition                                                                                                & \cmark         & \cmark       & \xmark        \\ \hline

~\cite{mathur2017deepeye}               & Deepeye                  & \begin{tabular}[c]{@{}l@{}}To perform image recognition with significant\\  accuracy in short duration\end{tabular}                  & \begin{tabular}[c]{@{}l@{}}Optimizing the operation of convolutional \\ and fully connected layers\end{tabular}                      & \cmark         & \cmark       & \xmark        \\ \hline

~\cite{shen2019deepapp}                 & DeepApp                                                                               & \begin{tabular}[c]{@{}l@{}}Determines the mobile device usage \\ pattern using historical data\end{tabular}                                & Deep reinforcement learning framework                                                                                     & \cmark         & \cmark       & \cmark        \\ \hline

~\cite{xue2019deepfusion}               & DeepFusion                                                                            & Recognition of different IoT based task            & \begin{tabular}[c]{@{}l@{}}A unified framework that incorporates values \\ of multiple sensors\end{tabular}               & \cmark         & \cmark       & \cmark                \\ \hline

~\cite{svyatkovskiy2017training}        & \textemdash   & \begin{tabular}[c]{@{}l@{}}To reduce the resource requirement through\\ distributed operations\end{tabular} &  \begin{tabular}[c]{@{}l@{}}Implemented a technique to train the DNN\\  model in a distributed manner\end{tabular}                                                                   &  \cmark             &  \cmark           &   \cmark           \\ \hline

~\cite{wang2018deep}                    & \textemdash    &  \begin{tabular}[c]{@{}l@{}}Issues associated with DNN model running \\on high-end machine and mobile devices\end{tabular}   & \begin{tabular}[c]{@{}l@{}}Elaborates the contradiction between \\ requirement of DNN model and mobile devices\end{tabular}                                                                                                                                                              &  \cmark             &   \cmark          &  \cmark            \\ \hline
~\cite{zhang2018shufflenet}             & \multicolumn{1}{l|}{Shufflenet}     &   \begin{tabular}[c]{@{}l@{}} To reduce the power consumption by reducing\\ the floating-point operations without accuracy \end{tabular}   &   \begin{tabular}[c]{@{}l@{}}Use pointwise group convolution  and  \\ channel shuffle \end{tabular}                                                                                                                                                                            &  \cmark             &  \xmark           &   \xmark           \\ \hline
\end{tabular}
}
\label{table6}
\end{table*}

Authors in~\cite{yao2017deepsense} proposed a unified framework that can handle the noise insertion in the collected dataset due to on-device sensors. Since the exact estimation of the noise distribution is highly tedious; therefore, a classification approach is needed that can automatically handle the noise in the dataset. Along with noise handling, the on-device classification is also beneficial to reduce the inherent delay due to data transmission from the device to the high-end machine. To solve these problems, simultaneously, the authors proposed a mechanism named DeepSense. DeepSense uses CNN and RNN based feature extraction to avoid the difficulty in designing manual features. As it is not always convenient to find highly robust features that can accommodate noise in the sensory data and rapidly changing user behaviour. DeepSense uses parallel execution of multiple sensors data on DNN models that decrease its complexity to run on mobile devices.

Yao \textit{et. al.}~\cite{yao2018fastdeepiot} proposed a framework that provides an insight into the black box of DNN model. The framework is named as FastDeepIoT that provides an accuracy preserving DNN compression for IoT devices. FastDeepIoT exploits the non-linear relation between the structure of DNN model and its execution time. Using the non-linear relation, the authors determined the tradeoff between execution time and DNN accuracy on the embedded device. Authors perform a high level of device profiling for automatically determining the minimal execution time on an embedded device with little accuracy compromise. The objective behind the FastDeepIoT is to develop a DNN model that check the conditions which are responsible for non-linearity without any knowledge of library and hardware.

In~\cite{abc} authors explored the tradeoff between performance and resources of the DNN model on the embedded device. Here, the authors determine the user-specified needs to estimate the appropriate DNN model for its deployment on the embedded device. The authors referred to this framework as AdaDeep. The framework automatically specifies a combination of compression techniques for a given DNN model. This combination of compression techniques leads to an optimal balance between resources and user demanded performance. It imposes resource constraint on accuracy, energy consumption, storage, and latency. AdaDeep relies on reinforcement learning-based optimization that automatically and efficiently solves the constraints optimization problem. The AdaDeep tried to solve following optimization problem
\begin{align}
 \arg \max_{c_t \in C_t} \text{  } & \lambda_1 \tau(acc-acc_{min}) + \lambda_2 \tau(E_{max}-E),\\ 
 \text{s.t.,  } & L \leq L_{th},\\ 
 & St \leq St_{th},
\end{align}

where, accuracy ($acc$), energy ($E$), latency ($L$), and storage ($St$) have the corresponding threshold $acc_{min}$, $E_{max}$, $L_{th}$, and $St_{th}$, respectively. The accuracy and energy constants are combined using constraints $\lambda_1$ and $\lambda_2$. $\tau(\cdot)$ is the normalization process for both accuracy and energy.

Radu \textit{et. al.}~\cite{radu2016towards} mainly focus on determining the capability of a DNN model to perform recognition of sensory values on mobile and embedded devices. The authors utilize cross-sensor correlation for providing a high-order of recognition performance. In~\cite{fang2018nestdnn} authors presented a DNN framework named NestDNN. The framework emphasized on the availability of dynamic resources on mobile devices. It indicates the DNN model running on a mobile device faces resource scarcity when different applications are running on the mobile device. Therefore, it is beneficial to select optimal resources while performing inference on the mobile device. NestDNN has the main objective of providing on-device computation of DNN model without Cloud support. NestDNN uses DNN model pruning and recovery method that transforms the cumbersome teacher model to a compact multi capacity student model.

Authors in~\cite{radu2018multimodal} studied the different benefits of using multimodal-sensors for recognizing various activities using DNN classifier on embedded devices. They mainly emphasized on two variant of DNN model, \textit{i.e,} fully connected layer and a convolutional neural network that is using the concatenation of multi-sensors values. Here, the authors have used the modality-specific partition. Mathur \textit{et. al.}~\cite{mathur2017deepeye} presented a DNN compression technique called DeepEye. The DeepEye provide a matchbox size computational unit having attached camera sensor. The small size device can process the image captured by the camera in the same way the image processed at high-end machines (Cloud). On the matchbox size device, the DNN model runs successfully to perform image recognition with significant accuracy in short duration. The local execution is possible by optimizing the operation of convolutional and fully connected layers in the DNN model. The convolutional layers are computationally expensive but require low storage. On the other hand, fully connected layers are computationally inexpensive and require higher storage than convolution layer. This reciprocal relationship between convolutional and fully connected layer creates a tradeoff for optimizing a DNN model. If the embedded device has sufficient storage and limited processing capacity, then perform the maximum operations by loading the fully connected layer on the device. Similarly, in vice-versa condition, the convolutional layers are loaded on the device.

Authors in~\cite{xu2018deepcache} proposed a cache designing mechanism that helps in providing inference on mobile devices. The authors named the approach as DeepCache. They suggest that the cache memory must be available and should handle the memory variation in the raw sensory data. The caching provide space to store results and helps in exploiting result reusability. The authors claim to require minimal efforts of the developer in designing the proposed mechanism. Through experimental analysis DeepCache have proved to reduce execution time by $18\%$ and energy consumption by $20\%$. In~\cite{shen2019deepapp} authors proposed a deep reinforcement learning framework named DeepApp. It determines the mobile device usage pattern using historical data of the applications being used by the user.  The features from the historical data are extracted using compressed DNN model. Further, the compressed DNN classifier recognizes the different usage pattern.

Xue \textit{et. al.}~\cite{xue2019deepfusion} proposed a unified framework (named DeepFusion) that incorporates values of multiple sensors for recognition of different IoT based task. DeepFusion parallelly combines spatial features extracted from CNN for each sensory values. This parallelisation reduces the complexity of the DNN model and makes it suitable for its deployment on mobile devices. The multimodal sensory value is considered as it not only provides better distinguishable characteristics but also reduces the noise in the sensory data. In~\cite{svyatkovskiy2017training} authors have implemented a technique to train the DNN model in a distributed manner. Here, the author trends to perform the execution on multiple GPUs. The authors tried to reduce the resource requirement through distributed operations.

Authors in~\cite{wang2018deep} elaborates the contradiction between resource requirements of the DNN models and deployment of DNN model on miniature mobile devices having limited resources. Other issues associated with training of DNN models on the high-end machine is the privacy and security concern about individual data. The cumbersome DNN models exceed the limited storage on the mobile device and its deployment off the device results in higher energy consumption and delay. It is tedious to perform both training and inference on mobile devices. During the training of DNN model on mobile devices, it requires distributed data generation and processing. Further, upon running a DNN model, it can dominate the entire processing on the device due to a huge number of floating-point operations. In~\cite{zhang2018shufflenet} authors have incorporated two crucial operations, \textit{i.e.,} pointwise group convolution and channel shuffle for compressing DNN model. The proposed approach was named as ShuffleNet that reduces the power consumption by reducing the floating-point operations without accuracy compromise.

\noindent $\bullet$\textbf{Remarks:} The DNN compression techniques, DeepSense~\cite{yao2017deepsense}, FastDeepIoT~\cite{yao2018fastdeepiot}, AdaDeep~\cite{abc}, NestDNN~\cite{fang2018nestdnn}, DeepEye~\cite{mathur2017deepeye}, DeepCache~\cite{xu2018deepcache}, DeepApp~\cite{shen2019deepapp}
DeepFusion~\cite{xue2019deepfusion}, ShuffleNet~\cite{zhang2018shufflenet}, \cite{radu2018multimodal,svyatkovskiy2017training,wang2018deep} in miscellaneous approach have reduced the storage and computation requirements with a different perspective. For example, DeepSense proposed a lightweight feature extraction mechanism using parallelization of sensory values and DeepFusion combines spatial and temporal features of sensory values using a parallelized architecture. Through, these mechanisms are effective and shown significant performance on smartphone using compressed DNN models. However, a higher degree of compression is needed for RCDs, where, storage and computation resources are much smaller than that of smartphones.

\vspace{0.8cm}

\section{Discussions and future research directions}\label{fd}
The presented categorization for compressing DNN model gives a glimpse of the existing literature and their colossal contributions. After a thorough review of the existing literature on DNN compression, we come up with five broad categories, \textit{i.e.,} network pruning, sparse representation, bits precision, knowledge distillation, and miscellaneous techniques. All these categories seem to be the predominating area of research that encourages effective utilization of DNN model on resource constraint devices. As, demand for IoT based applications is proliferating,  which enables data scientists to incorporate DNN models in IoT. Therefore, it is beneficial to provide a thorough overview of the DNN compression technique that can meet out the limited storage and processing capacity available at the resource constraint IoT devices.

The overview of different categories of DNN compression provides several research possibilities. Even though several work have been done in the DNN compression, there are still a lot more to be uncovered. The existing literature mainly emphasized on compressing convolutional neural network and fully connected layers, but very few work have been done for the recurrent neural networks. Additionally, despite different accuracy preserving mechanism adopted in existing literature, still, the DNN compression techniques face performance degradation. Finally, the dynamic resources on tiny computing devices need more exploration.

\begin{table}[htbp!]
\section*{Abbreviations and symbols}
\vspace{0.2cm}
\begin{tabular}{llll}
BNN         &  &  & Binarized Neural Network                          \\
CNN         &  &  & Convolutional Neural Network                      \\
cDeepArch   &   &   & compact Deep neural network Architecture        \\
DeepIoT     &  &  & Deep neural network for IoT                       \\  
DNN         &  &  & Deep Neural Network                               \\ 
EIE         &  &  & Energy efficient Inference Engine                 \\
EvoNAS      &  &  & Evolutionary Network Architecture Search          \\
FastDeepIoT &  &  & Fast Deep neural network for IoT                  \\
FLOPs       &  &  & FLoating point OPerations                         \\
FSKD        &  &  & Few Sample Knowledge Distillation                 \\
GPU         &  &  & Graphical Processing Unit                         \\
IoT         &  &  & Internet of Things                                \\
LSTM        &  &  & Long Short Term Memory                            \\
MobileDA    &  &  & Mobile Domain Adaptation                          \\
MTZ         &  &  & Multi Tasking Zipping                             \\
QNN         &  &  & Quantized Neural Network                          \\
RCD         &  &  & Resource Constraint Device                        \\
RNN         &  &  & Recurrent Neural Network                          \\
RONA        &  &  & pRivate mOdel compressioN frAmework               \\
SCNN        &  &  & Sparse Convolutional Neural Network               \\
SVD         &  &  & Singular Value Decomposition                      \\
VarGNet     &  &  & Variable Group convolution in DNN \\
%KL divergence &  &  &   Kullback Liebler divergence                                              \\
$\mathcal{D}$ &  &  &    Dataset                                               \\
$N$            &  &  &   Number of data instances                                                \\
$M$            &  &  &  Length of data instance                                                 \\
$\mathcal{L}(\cdot)$            &  &  & Loss function                                                  \\
$\mathbf{F}$            &  &  & Convolutional filter                                                   \\
$\mathbf{I}$            &  &  &   Input dimension                                                \\
$\mathbf{O}$            &  &  &  Output dimension                                                   \\
$\mathbf{W}$            &  &  &   Weight matrix                                                \\
$l$            &  &  &  Number of classes in $l$                                                 \\
$Clip(\cdot)$            &  &  &  Clip function                                                  \\
$Sign(\cdot)$            &  &  &  Sign function                                                 \\
$\mathcal{L}_{adl}(\cdot)$            &  &  &   Angular distillation loss                                                \\
$trans(\cdot)$            &  &  &   Transformation function                                                \\
$\mathbf{X}$            &  &  & Input matrix                                                  \\
$\mathbf{L}_{comb}$            &  &  &    Combined loss                                               \\

\end{tabular}
\end{table}

\bibliographystyle{IEEEtran}
\bibliography{btp}

\end{document}